\documentclass[lettersize,journal]{IEEEtran}
\usepackage{amssymb}
\usepackage{pifont}
\newcommand{\xmark}{\ding{55}}%
\usepackage[dvipsnames, svgnames, x11names]{xcolor}
\usepackage{colortbl}
\usepackage{times}
\usepackage{epsfig}
\usepackage{graphicx}
\usepackage{bm}
\usepackage{booktabs}
\usepackage{setspace}
\usepackage{float}
\usepackage{color}
\usepackage{multirow}
\usepackage{algorithm}

\usepackage[caption=false,font=normalsize,labelfont=sf,textfont=sf]{subfig}
\usepackage{textcomp}
\usepackage{stfloats}

\usepackage{algorithmic}  
\usepackage{amsfonts} 
\usepackage{listings}
\usepackage{ amsmath,  overpic, textpos}
\usepackage{stackengine}

\usepackage{multirow}
\usepackage{arydshln}
\usepackage{amsthm}

\usepackage{nicefrac}       
\usepackage{microtype}      
\usepackage{makecell}       
\usepackage{float}          

\usepackage{verbatim}
\usepackage{color}
\usepackage{float}
\usepackage{enumitem}
\usepackage{booktabs}
\usepackage[numbers,sort&compress]{natbib}
\usepackage{bbm}

\usepackage[pagebackref=false, breaklinks=true, letterpaper=true, colorlinks,
            citecolor=citecolor, linkcolor=linkcolor, bookmarks=false]{hyperref}
\definecolor{citecolor}{HTML}{0071BC}
\definecolor{linkcolor}{HTML}{ED1C24}
\definecolor{Ins_enc}{HTML}{4672C4}
\definecolor{mygray}{gray}{.9}

\newcommand{\gray}[1]{\textcolor{gray}{{#1}}}

\newlength\savewidth
\newcommand{\tablestyle}[2]{\setlength{\tabcolsep}{#1}\renewcommand{\arraystretch}{#2}\centering\footnotesize}
\renewcommand{\paragraph}[1]{\vspace{1.25mm}\noindent\textbf{#1}}

\newcolumntype{x}[1]{>{\centering\arraybackslash}p{#1pt}}
\newcolumntype{y}[1]{>{\raggedright\arraybackslash}p{#1pt}}
\newcolumntype{z}[1]{>{\raggedleft\arraybackslash}p{#1pt}}

\newcommand{\app}{\raise.17ex\hbox{$\scriptstyle\sim$}}

\definecolor{deemph}{gray}{0.6}

\definecolor{baselinecolor}{gray}{.9}
\definecolor{deltacolor}{gray}{.45}
\newcommand{\baseline}[1]{\cellcolor{baselinecolor}{#1}}

\definecolor{airforceblue}{rgb}{1.00, 0.501, 0.01}
\definecolor{egreen}{rgb}{0, 0.69, 0.314}

\definecolor{cyan}{cmyk}{.3,0,0,0}
\usepackage[capitalize]{cleveref}
\crefname{section}{Sec.}{Secs.}
\Crefname{section}{Section}{Sections}
\Crefname{table}{Table}{Tables}
\crefname{table}{Tab.}{Tabs.}

\usepackage{soul}
	\definecolor{airforceblue}{rgb}{0.36, 0.54, 0.66}

\newcommand{\modelname}{LMM-VQA}

\begin{document}
	
\title{LMM-VQA: Advancing Video Quality Assessment with Large Multimodal Models}
\author{
            Qihang Ge, Wei Sun, Yu Zhang, Yunhao Li, 
            Zhongpeng Ji, Fengyu Sun, \\
            Shangling Jui, Xiongkuo Min,~\IEEEmembership{Member, IEEE}, \\
	        Guangtao Zhai,~\IEEEmembership{Senior Member, IEEE}
\thanks{
Q. Ge, W. Sun, Y. Li, X. Min and G. Zhai are with the Institute of Image Communication and Information Processing, Shanghai Jiao Tong University, Shanghai 200240, China (e-mail:\{qihang.ge, sunguwei, lyhsjtu, minxiongkuo, zhaiguangtao\}@sjtu.edu.cn)
}
\thanks{
Y. Zhang is with the MoE Key Laboratory of Artificial Intelligence, AI Institute, Shanghai Jiao Tong University, Shanghai 200240, China (e-mail: cynthiazhang@sjtu.edu.cn)
}
\thanks{
Z. Ji, F. Sun, and S. Jui are with the Huawei Technologies, Shanghai 200240, China. (e-mail: jizhongpeng@huawei.com; sunfengyu@hisilicon.com; jui.shangling@huawei.com)
} 
}
\maketitle
\begin{abstract}


The explosive growth of videos on streaming media platforms has underscored the urgent need for effective video quality assessment (VQA) algorithms to monitor and perceptually optimize the quality of streaming videos. However, VQA remains an extremely challenging task due to the diverse video content and the complex spatial and temporal distortions, thus necessitating more advanced methods to address these issues. Nowadays, large multimodal models (LMMs), such as GPT-4V, have exhibited strong capabilities for various visual understanding tasks, motivating us to leverage the powerful multimodal representation ability of LMMs to solve the VQA task.
Therefore, we propose the first \textit{L}arge \textit{M}ulti-\textit{M}odal based \textit{V}ideo \textit{Q}uality \textit{A}ssessment (LMM-VQA) model, which introduces a novel spatiotemporal visual modeling strategy for quality-aware feature extraction. Specifically, we first reformulate the quality regression problem into a question and answering (Q\&A) task and construct Q\&A prompts for VQA instruction tuning. Then, we design a spatiotemporal vision encoder to extract spatial and temporal features to represent the quality characteristics of videos, which are subsequently mapped into the language space by the spatiotemporal projector for modality alignment. Finally, the aligned visual tokens and the quality-inquired text tokens are aggregated as inputs for the large language model (LLM) to generate the quality score as well as the quality level. 
Extensive experiments demonstrate that \textit{\modelname{}} achieves state-of-the-art performance across five VQA benchmarks, exhibiting an average improvement of $5\%$ in generalization ability over existing methods. Furthermore, due to the advanced design of the spatiotemporal encoder and projector, \modelname{} also performs exceptionally well on general video understanding tasks, further validating its effectiveness. Our code will be released at \url{https://github.com/Sueqk/LMM-VQA}.

\end{abstract}
	

\begin{figure}[t]
    \centering
    \includegraphics[width=\linewidth]{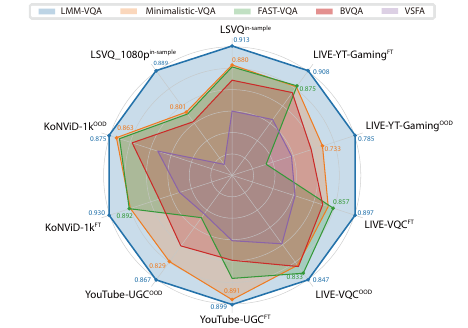}
    \captionsetup{size=small}
    \caption{An overall performance comparison between \textbf{\textsc{\modelname{}}} and existing state-of-the-arts methods. KoNViD-1k\(^\text{OOD}\) indicates that the model is trained on the LSVQ dataset and evaluated on the KoNViD-1k dataset. KoNViD-1k\(^\text{FT}\) refers that the model is pre-trained on the LSVQ dataset, fine-tuned on KoNViD-1k, and evaluated on the KoNViD-1k dataset. LSVQ\(^\text{in-sample}\) represents that the model is trained on the LSVQ dataset and evaluated on the LSVQ dataset. Metrics are (SRCC+PLCC)/2.}
    \label{fig:radar}
        \vspace{-5pt}
\end{figure}


\section{Introduction}
\label{sec:intro}

\IEEEPARstart{A}{s} video consumption continues to surge across a variety of streaming media platforms, including YouTube, TikTok, Facebook, Netflix, etc, accurately evaluating the perceptual quality of videos is crucial for these video-enabled applications and services to monitor video quality throughout the entire video processing procedures and perceptually optimize video procession algorithms (\textit{e.g.} video compression and enhancement), ultimately providing better quality of experience (QoE) for end-users. Towards this goal, several video quality assessment (VQA) methods~\cite{min2024perceptual} have been proposed in recent years, which can be categorized into reference-based VQA, including full-reference~\cite{bampis2018spatiotemporal, sun2021deep} and reduced-reference VQA~\cite{gunawan2008reduced, soundararajan2012video}, and no-reference VQA, also known as blind VQA (BVQA)~\cite{saad2014blind,Korhonen2019TLVQM,li2019quality,tu2021rapique,tu2021VIDEVAL,yi2021attention,lsvq,wang2021rich,li2022bvqa,simple-vqa,fast-vqa,zhang2023md,wu2023dover,sun2024enhancing,sun2024analysis}. In this paper, we focus on BVQA\footnote{In the following paper, we also use VQA to refer to BVQA for simplicity.} as it does not require any reference video information and has a broader range of application scenarios.

In the literature, BVQA is typically studied from two perspectives: the knowledge-driven approach~\cite{mittal2012no,Korhonen2019TLVQM,tu2021rapique,tu2021VIDEVAL} and the data-driven approach~\cite{VSFA,lsvq,wang2021rich,li2022bvqa,simple-vqa,fast-vqa,wu2023dover}. Knowledge-driven BVQA methods leverage prior knowledge of visual quality perception to extract corresponding handcrafted features for quality regression. For example, natural scene statistics (NSS)~\cite{mittal2012making} have been proven to be sensitive to synthetic distortions and are the most frequently used features in knowledge-driven blind image and video quality assessment studies~\cite{mittal2012no,saad2012blind,mittal2015completely}. Other visual descriptors, such as texture~\cite{ding2020image}, noise~\cite{ghadiyaram2017capture}, contrast~\cite{gu2015analysis}, blockiness~\cite{zhu2014no}, color~\cite{shang2018color}, motion vector~\cite{chen2011prediction}, optical flow~\cite{manasa2016optical}, etc., are also considered quality-related features and have been explored in previous studies. 


\begin{table*}[ht]
\centering
\small
\renewcommand\arraystretch{1.18}
\vspace{-5pt}

\caption{The performance of latest LMMs on varying VQA datasets. We utilize the text prompts and video as input and ask LMMs to generate the specific scores.}
\setlength{\tabcolsep}{1.7mm}{\begin{tabular}{l:cc:cc:cc:cc:cc:cc}

\hline
\multicolumn{1}{l}{\textbf{Dataset}} & \multicolumn{2}{c}{\textbf{LSVQ\(_\text{test}\)}} & \multicolumn{2}{c}{\textbf{LSVQ\(_\text{1080p}\)}} & \multicolumn{2}{c}{\textbf{KoNViD-1k}} & \multicolumn{2}{c}{\textbf{YouTube-UGC}} & \multicolumn{2}{c}{\textbf{LIVE-VQC}}& \multicolumn{2}{c}{\textbf{LIVE-YT-Gaming}}\\ 
\hline
\multicolumn{1}{l:}{\textbf{Method}}        &\multicolumn{1}{c}{SRCC}  &\multicolumn{1}{c:}{PLCC}   &\multicolumn{1}{c}{SRCC} &\multicolumn{1}{c:}{PLCC}   &\multicolumn{1}{c}{SRCC}&\multicolumn{1}{c:}{PLCC}&\multicolumn{1}{c}{SRCC}&\multicolumn{1}{c}{PLCC}&\multicolumn{1}{c}{SRCC}&\multicolumn{1}{c}{PLCC}&\multicolumn{1}{c}{SRCC}&\multicolumn{1}{c}{PLCC}  \\
\hline 
Video-ChatGPT~\cite{maaz2023video}                 
&0.060&0.055	&0.055&0.065	&0.010&0.058	&0.037&0.054	&0.005&0.008	 &0.051&0.050  \\
LLaMA-VID~\cite{li2023llama}                 
&0.121&0.133	&0.093&0.119	&0.145&0.183	&0.091&0.118	&0.099&0.125	&0.023&0.045  \\
VideoLLaMA2~\cite{cheng2024videollama2}  
&0.220&0.294	&0.129&0.227	&0.080&0.142	&0.174&0.260	&0.090&0.124	&0.007&0.058  \\
LLaVA-NeXT-Video~\cite{li2024llava}                 &0.288&0.318    &0.246&0.274  &0.287&0.381  &0.287&0.381 &0.094&0.128&0.171&0.175  \\
VILA-1.5~\cite{lin2024vila}                 
&0.470&0.484	&0.432&0.508	&0.566&0.581	&0.412&0.465	&0.392&0.461	&0.217&0.243  \\

\hline

\end{tabular}}
\label{tab:LMM}
\vspace{-5pt}
\end{table*}


In contrast, data-driven BVQA methods automatically learn the quality-aware feature representation by training a carefully designed deep neural network (DNN) in a learning-based manner. Typically, there are three kinds of architectures for quality-aware feature extraction: 1) utilizing a 2D network to extract frame-wise features and then employing a sequence model (\textit{e.g.} GRU~\cite{cho2014gru}) to fuse them into video-level features~\cite{VSFA,yi2021attention}; 2) utilizing a 2D network to extract key frame features and a 3D network to extract the motion features from video chunks, which are then fused into video-level features by the concatenating operator or a sequence model~\cite{simple-vqa,wu2023dover}; 3) directly utilizing a 3D network to extract video-level features from video chunks~\cite{fast-vqa}. The 2D and 3D backbones can be pre-trained from other computer vision tasks~\cite{Ying2020PaQ-2-PiQ,he2016deep} or fine-tuned on VQA datasets via an end-to-end manner.

Despite that significant efforts have been made in the field of BVQA, existing methods still face several challenges. First, knowledge-driven BVQA methods have better explainability but perform poorly on in-the-wild videos, as handcrafted features cannot model the complex in-the-wild distortion diverse video content. 
Second, although data-driven BVQA methods perform well on specific VQA datasets, they struggle to maintain the same level of performance during out-of-distribution (OOD) evaluations (as shown in Fig.~\ref{fig:radar}), which is more critical for real-world applications. This limitation arises because current VQA datasets contain a relatively small scale of videos, which are insufficient to cover the vast high-dimensional spaces of video content and distortions, thereby restricting the ability of trained BVQA models to generalize to OOD videos.

To overcome this problem, a straightforward approach is to perform subjective experiments to construct a larger-scale VQA dataset that encompasses a wider variety of video samples. However, subjective VQA experiments are highly expensive and time-consuming, making the construction of such a large-scale VQA dataset challenging. Another routine is to leverage as many existing visual understanding models as possible to enhance the feature representation of BVQA models. For example, some BVQA studies~\cite{wang2021rich,yuan2024ptm,liu2023ada,sun2024enhancing} combine diverse visual task models and quality assessment models to boost their representation capacity. Recently, the emergence of large multimodal models (LMMs)~\cite{achiam2023gpt,ye2023mplug,liu2024visual} opens the door for vision foundation models to handle general-purpose visual tasks. Some work also reveals that LMMs have strong abilities to understand low-level visual characteristics~\cite{zhang2023internlm, wu2023q-instruct, liu2023LLaVA-1.5, wu2023q-bench}. These encouraging advances inspire us to leverage LMMs to address the BVQA task, thus benefiting from their strong general-purpose visual representation capabilities. However, our empirical studies reveal that directly utilizing video LMMs as the quality evaluator lead to very poor performance. As shown in Table~\ref{tab:LMM}, the SRCC and PLCC values of the best-performing model, VILA-1.5, do not exceed $0.6$ on five VQA benchmarks. Therefore, \textit{how to adapt LMMs to the VQA task and leverage their strong visual representation capabilities to develop effective and robust VQA model remains an open challenge}.

\begin{figure*}[ht]
    \centering
    \vspace{-5pt}
    \includegraphics[width=\linewidth]{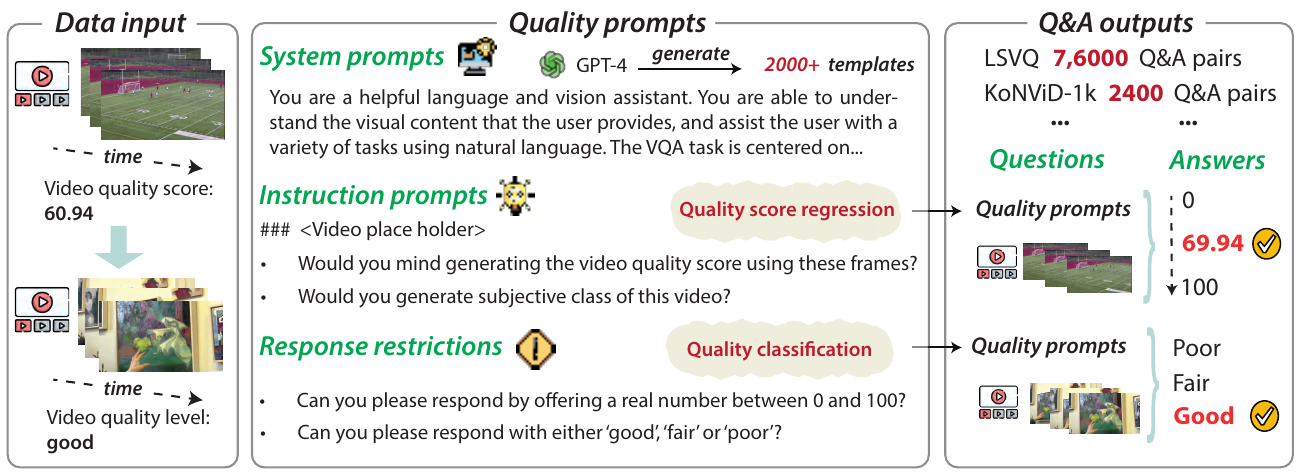}
    \captionsetup{size=small}
    \caption{\textbf{ The construction of pair-wised Q\&A instruction prompts for the training of \modelname{}}. Given the video input, and its quality score, we construct pair-wised Q\&A instruction prompts for two tasks: quality score regression and quality classification. We leverage the GPT-4 annotator to generate 2000 templates of quality prompts, which include system prompts, instruction prompts, and response restrictions. }
    \label{fig:prompt}
    \vspace{-5pt}
\end{figure*}

Towards this goal, we propose \textbf{the first large multimodal-based video quality assessment (LMM-VQA) model} by instruction tuning a spatiotemporal enhanced LMM model on the VQA datasets. Given that the functionality of LMMs is defined by instruction prompts, we design a rule-based approach to automatically generate a set of question-and-answer (Q\&A) pairs for the VQA datasets. As shown in Fig.~\ref{fig:prompt}, the question template, termed quality prompts, consists of one system prompt, two instruction prompts, and two response restrictions to ensure the LMM comprehends the VQA task and produces the correct output format. The answering template contains the video quality score and the quality level, serving as the ground truth outputs for fine-tuning the LMM-VQA model.

Our LMM-VQA model consists of three modules: a spatiotemporal enhanced visual encoder for video feature extraction, a spatiotemporal visual project for vision-language alignment, and a large language model (LLM) for understanding video quality characteristics. For the spatiotemporal enhanced visual encoder, we avoid the sparse sampling frame strategy commonly used in most video LMMs~\cite{ye2023mplug,lin2023video-llava,li2024videochat} to encode video features, as temporal distortions in videos are typically reflected in continuous frames. Instead, we treat the video features as two distinct parts: spatial features and temporal features, where the spatial features are extracted by a 2D visual encoder (\textit{e.g.,} CLIP ViT-L/14~\cite{radford2021learning}) from sparse key frames, while the temporal features are extracted by a 3D visual encoder (\textit{e.g.,} SlowFast~\cite{SlowFast}) from continuous frames. This approach empowers LMMs to capture enough spatial and temporal distortion cues in videos for quality analysis, thereby enhancing prediction performance for the VQA task. 
Subsequently, the spatial and temporal features are inputted to the spatiotemporal projector to map visual tokens into the language space for modality alignment. Finally, the aligned visual tokens and quality prompt tokens are aggregated as input for the LLM to generate the final answers, including the video quality scores and levels. Experimental results reveal that \modelname{} achieves state-of-the-art (SOTA) performance across different datasets. Specifically, we apply in-sample, OOD, and fine-tuning evaluations to demonstrate its effectiveness in VQA tasks.


Our contributions are summarized as follows:

\begin{itemize}
    \item We develop an instruction tuning pipeline to adapt LMMs for the VQA task, effectively leveraging their powerful visual representation capabilities for high-performance and robust video quality evaluation.
    \item To address tricky temporal distortion issues in videos, we propose a spatiotemporal enhanced visual encoder that extracts the spatial and temporal features separately. The temporal features enhance the perception of temporal distortions and significantly improve the performance of the proposed \modelname{}.
    \item LMM-VQA achieves superior performance on five VQA benchmarks in both fine-tuning and OOD evaluation settings. What's more, LMM-VQA also perform well on general video understanding benchmarks. These experimental results comprehensively demonstrate the effectiveness of LMM-VQA.
    
\end{itemize}

\section{Related Work}
\label{sec:re}

\subsection{Blind Video Quality Assessment}

\begin{figure}[ht]
    \vspace{-5pt}
    \centering
    \includegraphics[width=\linewidth]{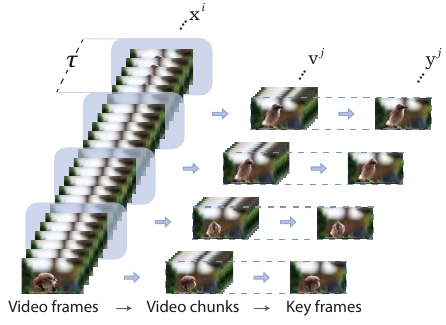}
    \captionsetup{size=small}
    \caption{\textbf{Illustration of the video preprocessor}. 
We slice the video into non-overlapping consecutive chunks every $\tau$ frames. The first frame of $j$-th chunk $\mathbf v^j$ is selected as key frame $\mathbf y^j$.}
    \label{fig:chunk}
    \vspace{-5pt}
\end{figure}

\textbf{Knowledge-driven BVQA models} primarily utilize handcrafted features derived through prior knowledge of visual quality perception to evaluate video quality~\cite{mittal2012no, Korhonen2019TLVQM,tu2021rapique,tu2021VIDEVAL}. For instance, NIQE~\cite{mittal2012making} and BRISQUE~\cite{mittal2012no} first introduce natural scene statistics (NSS) for general quality predictions. Subsequently, V-BLIINDS~\cite{saad2012blind} and VIIDEO~\cite{mittal2015completely} extend NSS to the temporal domain to address the BVQA task. TLVQM~\cite{Korhonen2019TLVQM} integrates two-level features---spatial high-complexity and temporal low-complexity features---to measure complex spatial and temporal distortions. VIDEVAL~\cite{tu2021VIDEVAL} ensembles diverse handcraft features sourced from classical BI/VQA models to create a strong feature representation for VQA. In summary, knowledge-driven BVQA models typically incorporate temporal-related features into BIQA models or extend the 2D quality features into 3D, demonstrating the critical importance of temporal information in video quality evaluation. However, due to insufficient understanding of human perception in quality assessment, knowledge-driven BVQA models tend to underperform on in-the-wild videos.

\textbf{Data-driven BVQA models} have gained popularity with the advancement of DNNs, aiming to automatically learn quality-aware features from labeled video data~\cite{VSFA,lsvq,wang2021rich,li2022bvqa,simple-vqa,fast-vqa,wu2023dover}.For example, Li \textit{et al.}~\cite{VSFA} use GRU~\cite{cho2014gru} to capture the temporal relationship among the semantic features extracted by ResNet-50~\cite{he2016deep}. 
Liu \textit{et al.}~\cite{liu2018end} jointly optimize the feature extractor and the regressor for quality assessment and compression distortion classification. Wang \textit{et al.}~\cite{wang2021rich} utilizes the compression level, video content, and distortion type derived from independent networks to evaluate the quality of user-generated videos. Li \textit{et al.}~\cite{li2022bvqa} combine the CNN-based temporal model and RNN-based temporal model to improve the VQA task. Sun \textit{et al.}~\cite{simple-vqa} propose SimpleVQA, an efficient and effective BVQA framework that uses sparse high-resolution key frames to extract spatial features and dense low-resolution frames to extract motion features. This framework is further streamlined to Minimalistic BVQA~\cite{sun2024analysis}, which is used to analyze the characteristics of VQA datasets. Wu \textit{et al.}~\cite{fast-vqa} propose FAST-VQA, which samples video fragments (\textit{i.e.}, the spatially spliced and temporally aligned patches) as input to a modified video Swin Transformer~\cite{liu2022swin} to train an end-to-end BVQA model. They~\cite{wu2023dover} later ensemble an aesthetics perception branch to FAST-VQA to better extract spatial features. Although the methods achieve astonishing performance on video quality assessment tasks, they still suffer from compromised generalization abilities.

\subsection{Large Multimodal Models for Visual Quality Assessment}
The advent of large multimodal models, typified by GPT-4V, has triggered keen interest in leveraging such techniques to quality assessment tasks~\cite{zhang2024benchmark}. Recent studies have explored the potential of adopting LMMs for visual quality assessment~\cite{chen2023x-iqe, wu2023q-bench, wu2023q-instruct, q-align}. For the IQA task, Wu \textit{et al.}~\cite{wu2023q-bench} investigate the use of LMMs to predict quantifiable quality scores by extracting the softmax pooling result from the logits of two frequent tokens: `good' and `poor'. However, for tasks that require a precise visual quality understanding, LMMs without instruction fine-tuning have shown inferior performance compared to traditional methods~\cite{wu2023q-bench}. This is partly because LLMs struggle to produce accurate responses without a deep alignment between visual representation and scores. To address this, they proposed Q-instruct~\cite{wu2023q-instruct}, a fine-tuned LMM model that leverages instructed question-answering datasets focused on low-level visual scoring. Further, Wu \textit{et al.}~\cite{q-align} propose a human-like syllabus to teach LMMs using text-defined rating levels rather than generating scores directly. However, previous LMM methods only model intra-frame information through discrete frames, leading to a significant loss of temporal information.

\begin{figure*}[ht]
    \vspace{-5pt}
    \centering
    \includegraphics[width=\linewidth]{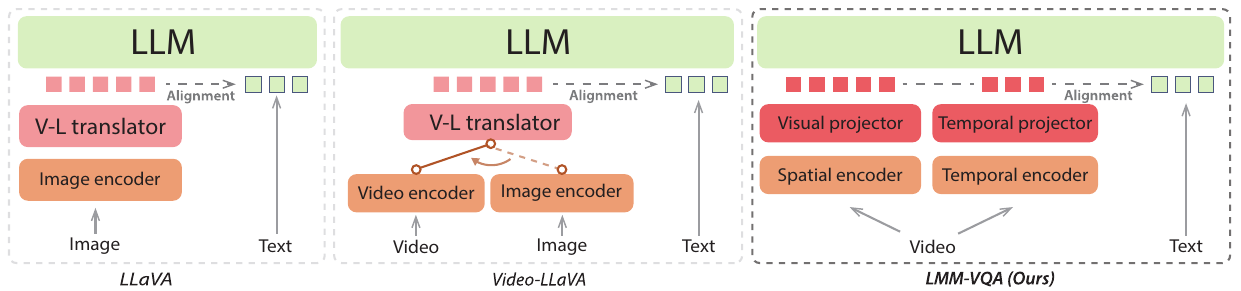}
    \captionsetup{size=small}
    \caption{\textbf{Technical comparisons with other methods}. \textbf{Left:} LLaVA takes the images as input only; \textbf{Middle:} the core architecture of Video-LLaVA. It shares a unified V-L translator for images and videos due to the small modality gap between the two modalities; \textbf{Right:} Our proposed method. We take two separate projectors for modality alignment to bridge the larger modality gap between video with temporal information and text.}
    \label{fig:model-llm}
    \vspace{-10pt}
\end{figure*}


\section{Methods}

\subsection{Problem Setting}
For an LMM-based video quality assessment model, we define $\mathbf X$ as the video input, $\mathbf T$ as the text input, and $\mathbf{ \hat{Q}}$ as the video quality, including quality scores or levels output. $\mathbf{ \hat{Q}}$ is an approximation of the ground-truth video quality $\mathbf Q$. Then, we define the assessment model $\mathcal{F}$ as follows:
\begin{align}
    \mathbf{\hat{Q}} = \mathcal{F}(\mathbf X, \mathbf T).
\end{align}

\subsection{Q\&A Instruction Prompts}
\label{ssec:data}
Large instruction-tuned multimodal models (\textit{i.e.}, fine-tuned to respond to instructions) showcase an impressive ability for new tasks. Instruction prompt generation is a method for inducing instruction following capabilities with minimal human labeled data~\cite{ouyang2022training}. Its effectiveness has been validated through extensive instruction tuning experiments~\cite{wang2022self}. In our work, we construct Q\&A instruction prompts using videos and their MOS, as illustrated in Fig.~\ref{fig:prompt}. The question part includes quality prompts and videos. Quality prompts can be divided into three parts: system prompts, instruction prompts, and response restrictions. System prompts delineate the role of the LMM in this task, ensuring that the LMM has a clear understanding of the task. Instruction prompts pose a direct question to the LMM, as a conversational distillation of the task description. Response restriction defines the desired format for the model's answers. A typical example of these prompts is shown in Fig.~\ref{fig:prompt}.

Specifically, we construct pair-wised Q\&A instruction prompts based on the VQA benchmark datasets~\cite{lsvq,konvid-1k,youtube-ugc,live-vqc,live-yt-gaming}. The basic analysis of these datasets is shown in Table~\ref{tab:dataset intro}. Note that each video-score sample in VQA datasets results in two Q\&A instruction prompts, separately for the quality score regression task and the quality classification task. 
For the quality score regression task, we take scores as the answer in Q\&A pairs. For the classification task, we equally divide dataset samples into three levels---\textit{'good'}, \textit{'fair'}, and \textit{'poor'}---based on their quality scores, and take these levels as the classification responses. To ensure the diversity of the training dataset, we generate $2,000$ templates of instructions by GPT-4 such as: ``Would you mind calculating the video quality score with the help of these frames?''. Meanwhile, the visual tokens \textlangle \texttt{image-i}\textrangle\; extracted from video chunks and key frames are inserted followed by quality prompts during our training.

\begin{figure*}[ht]
    \centering
    \includegraphics[width=\linewidth]{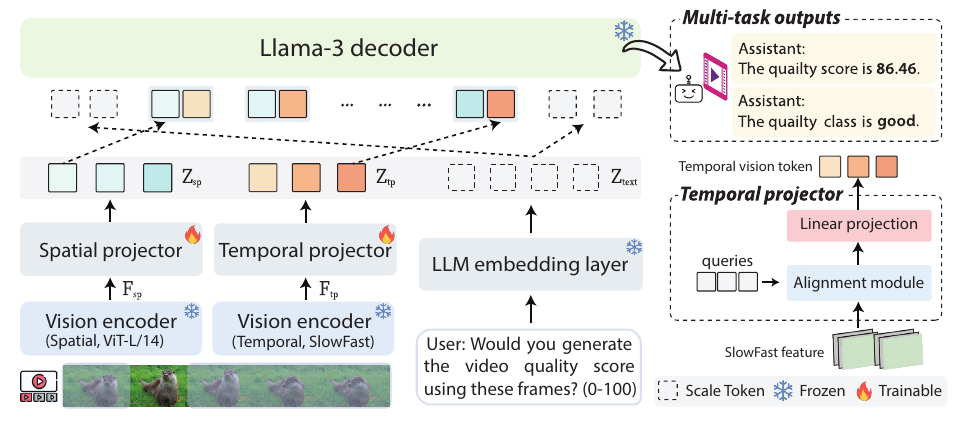}
    \captionsetup{size=small}
    \caption{\textbf{The framework of \modelname{}}: \modelname{} takes video frames as input, and generates responses of quality scores. The process initiates with two vision encoders that transform input frames into spatiotemporal visual features. These visual features are incorporated into two projection modules to generate visual tokens that are aligned with the text-related language space. Meanwhile, the text decoder produces scale tokens based on the quality prompt from users. Then, text-guided spatial and temporal tokens, and quality prompt tokens are aggregated as input for LLMs to generate the final answers.}
    \label{fig:overall}
    \vspace{-5pt}
\end{figure*} 

\subsection{Video Preprocessing}
To balance quality assessment efficacy and algorithmic complexity, we perform spatiotemporal downsampling operations during video preprocessing. Consider a video $\mathbf X = \{\mathbf x^i\}_{i=0}^{N-1}$, where $\mathbf x^i \in \mathbb{R}^{H\times W\times 3}$ represents the $i$-th frame with a resolution of $H\times W$, and $N$ is the total number of frames. 
As illustrated in Fig.~\ref{fig:chunk}, we aim to extract video chunks $\mathbf V \in \mathbb{R}^{K \times \tau \times H \times W \times 3}$ for temporal feature extraction and key frames $\mathbf Y \in \mathbb{R}^{K \times H \times W \times 3}$ for spatial feature extraction. 
Thus, we slice the video in the temporal domain to derive non-overlapping consecutive video chunks $\mathbf V = \{\mathbf v^j\}_{j=0}^{K-1}$. Each video chunk has the same frame length $\tau$ and $K = \lfloor N/\tau\rfloor$. $\lfloor \cdot\rfloor$ is the floor function. So we define the $j$-th video chunk $\mathbf v^j$ as:
\begin{align}
    \mathbf v^j = \{\mathbf x^i\}_{ i=\tau j}^{ \tau (j+1)},
\end{align}
Then, we select the first frame $\mathbf v^j$ as the $j$-th key frame $\mathbf y^j$:
\begin{align}
    \mathbf y^j = \mathbf x^{\tau j},
\end{align}
Subsequently, video chunks $V$ and key frames $Y$ are processed by the vision encoder and projector to obtain visual tokens that serve as input to the LLM decoder.

\subsection{Model Structure}


\begin{algorithm}[ht]
    \caption{\modelname{}} 
    \begin{algorithmic}[1]%
        \REQUIRE video chunks ${\mathbf V}$, key frames $\mathbf{Y}$, pair-wised instruction text prompts: ${\mathbf T}$
        \FOR{
        $\mathbf{y}^j $ in $\{\mathbf{y}^0, \mathbf{y}^1, \cdots, \mathbf{y}^{K-1}\}$ }
        \STATE $\mathbf{F}^j_{\mathbf{sp}} = f_{sp}(\mathbf{y}^j)$
        \ENDFOR
        \STATE All spatial vision embeddings: $\mathbf{F}_{\mathbf{sp}}\in\mathbb{R}^{N_p\times C_{sp}}$
        \FOR{$\texttt{\bf v}^j $ in  $\{{\bf v}^{0}, {\bf v}^{1}, \cdots, {\bf v}^{K-1} \}$}
        \STATE $\mathbf{F}^j_{\mathbf{tp}} = f_{tp}(\mathbf{v}^j)$
        \ENDFOR
        \STATE All temporal vision embeddings: $\mathbf{F}_{\mathbf{tp}}\in\mathbb{R}^{1\times C_{tp}}$
        \STATE Alignment for spatial token: 
            \\$\mathbf{Z_{sp}} = f_{ViT}\left(\mathbf{F_{sp}}\right)$
        \STATE Alignment for temporal token: 
            \\$\mathbf{Z_{tp}} = Mean\left(f_{MLP}\left(\mathbf{F_{tp}}\right)\right)$
        \STATE Aggregated input token: \\$\mathbf{Z_{all}} = f_{cat}\left(\mathbf{Z_{sp}}, \mathbf{Z_{tp}}, \mathbf{Z_{text}}\right)$
        \STATE Generate quality score or level by LLM decoder: \\$\mathbf{Z_{quality}} = \mathcal{D}\left(\mathbf{z}_{\ell}\mid\mathbf{Z_{all}}, {\mathbf{z}}_{<\ell}\right)$
        \ENSURE $\mathbf{Z_{quality}}$
    \end{algorithmic} 
    \label{alg:llm-vqa}
\end{algorithm}


Videos, comprised of sequences of frames, depict motion scenes that differ from static images~\cite{liu2022swin}. Therefore, the characteristics of both spatial and temporal dimensions are crucial for understanding videos~\cite{liu2023all}. Previous works have relied on discrete frames of videos to extract visual features, such as Video-ChatGPT~\cite{maaz2023video}. While discrete frames provide a fundamental basis for understanding videos, they do not make use of the temporal features of continuous frames, leading to overlooked issues such as jitter, lagging, flickering, etc. Thus, to address spatial and temporal distortion issues in videos, we propose a novel spatiotemporal enhanced encoder---a two-branch encoder designed to enhance the capture of spatial and temporal information in videos.

\begin{table*}[ht]
\vspace{-5pt}
\caption{Dataset statistics of five existing VQA datasets. We also show the data description of our proposed instruction Q\&A pairs. For each video and its quality score, we construct two Q\&A pairs for training.}
\small
\renewcommand\arraystretch{1.14}
\setlength{\tabcolsep}{2.9mm}{
    \begin{tabular}{c|cccc|cc}
      \hline
      \multirow{2}{*}{\textbf{Dataset}} &\multicolumn{4}{c|}{\bf Dataset Statistics} &\multicolumn{2}{c}{\bf Instruction Q\&A} \\
      \cline{2-7}
                     &\# of videos  &Resolution &Duration   &Source    &\# of pairs    & Annotator \\
      \hline
       LSVQ          &3.8k &99p-4K     &5-12s  & Internet Archive \& YFCC100M~\cite{thomee2016yfcc100m}  & 76k   & GPT-4 \\
      
       KoNViD-1k     &1.2k &540p       &8s     &YFCC100M~\cite{thomee2016yfcc100m}    & 2.4k     & GPT-4\\
       LIVE-VQC      &0.6k   &240p-1080p &10s    & Camera-captured                 & 1.2k       & GPT-4 \\
       YouTube-UGC   &1.5k  &360p-4K    &20s    &YouTube                                                & 3k        & GPT-4 \\
       LIVE-YT-Gaming&0.6k   &360p-1080p &8-9s   &Internet \& software-recorded                & 1.2k         & GPT-4 \\
     \hline
    \end{tabular}
}
 \label{tab:dataset intro}
 \vspace{-10pt}
\end{table*}

To detail the soundness of our technical design, we compare our \modelname{} with two similar approaches, LLaVA~\cite{liu2024visual} and Video-LLaVA~\cite{lin2023video-llava} respectively. As shown in Fig.~\ref{fig:model-llm}, LLaVA only takes the images as input without other external modalities. Different from LLaVA, Video-LLaVA takes images and discrete video frames as input and uses different vision encoders for them. With a small modality gap between images and discrete video frames, Video-LLaVA performs well with the shared V-L translator in video understanding tasks. In our task, temporal features are quite different from pixel-level video representation. Thus, we propose a spatiotemporal encoder with each independent projector to facilitate the modality alignment. 

Algorithm~\ref{alg:llm-vqa} illustrates the pipeline of \modelname{}, starting from key frames and video chunks. Specifically, as shown in, for the spatial branch, a 2D visual encoder (ViT-L/14~\cite{radford2021learning} in Fig.~\ref{fig:overall}) transforms the input key frames $\mathbf{Y}$ into spatial features $\mathbf{F_{sp}}$: 
\begin{align}
    \mathbf{F_{sp}} = f_{sp}(\mathbf{Y}), \mathbf{F_{sp}}\in\mathbb{R}^{K \times N_p\times C_{sp}},
\end{align}
where $ N_p= H/p\times  W/p$ indicate the number of image patches and and $C_{sp}$ is the number of spatial embedding dimensions, respectively. For the temporal branch, a 3D visual encoder (SlowFast in Fig.~\ref{fig:overall}) extracts temporal features $\mathbf{F_{tp}}$:
\begin{align}
    \mathbf{F_{tp}} = f_{tp}(\mathbf{V}), \mathbf{F_{tp}} \in \mathbb{R}^{K \times 1\times C_{tp}},
\end{align}
where $C_{tp}$ is the number of temporal embedding dimensions.


To address the disparity between vision and text modalities, 
we design a spatiotemporal projector to map visual tokens into language space for modality alignment, as illustrated in Fig.~\ref{fig:overall}. Specifically, we introduce a \textbf{spatial projector} which comprises a single-layer vision transformer (ViT) and a one-layer MLP, to map spatial features into the language space. Following this, we use a one-layer MLP as a \textbf{temporal projector} to align temporal features with the language space. We formulate these processes as:
\begin{align}
    \mathbf{Z_{sp}} = f_{ViT}\left(\mathbf{F_{sp}}\right),
\end{align}
\begin{align}
    \mathbf{Z_{tp}} = Mean\left(f_{MLP}\left(\mathbf{F_{tp}}\right)\right).
\end{align}

A text decoder generates scalable text tokens $\mathbf{F_{text}}$ from quality prompts $\mathbf{T}$ (\textit{i.e.}, LLM embedding layer in Fig.~\ref{fig:overall}). Spatial tokens $\mathbf{Z_{sp}}$, temporal tokens $\mathbf{Z_{tp}}$ and text tokens  $\mathbf{Z_{text}}$ are concatenated in the context length dimension $f_{cat}$, defined as $\mathbf{Z_{all}}$:
\begin{align}
    \mathbf{Z_{all}} = f_{cat}\left(\mathbf{Z_{sp}}, \mathbf{Z_{tp}}, \mathbf{Z_{text}}\right).
\end{align}
These text-guided visual tokens and quality prompt tokens are aggregated to generate the final output tokens, which represent the predicted quality scores or levels. 

We utilize the Llama-3 model~\cite{dubey2024llama3} as the language decoder and output the predicted answers in an auto-regressive manner:
\begin{align}
    \mathbf{Z_{quality}} = \mathcal{D}\left(\mathbf{z}_{\ell}\mid\mathbf{Z_{all}}, {\mathbf{z}}_{<\ell}\right),
\end{align}
where $\mathbf{Z_{quality}}=\{\mathbf z_l\}_{l=0}^{L-1}$ is the output text sequence follows the input tokens $\mathbf{Z_{all}}$ and $\mathcal{D}(\cdot)$ is the Llama-3 decoder. $\mathbf{z}_l \in \{0, 1\}^{|\mathbb{S}|}$ and $\mathbb{S}$ denote the vocabulary set. With well-trained \modelname{}, $\mathbf{Z_{quality}}$ represents an approximation $\mathbf{\hat{Q}}$ to the true perceptual quality $\mathbf{Q}$. The training process uses a cross-entropy loss: 
\begin{align}
    \mathcal{L}=-\sum_{\ell=0}^{L-1} \log \bm p\left({\mathbf{z}}_{\ell} \mid \mathbf{Z_{all}}, \mathbf{z}_{<\ell}\right).
\end{align}

We use pair-wised Q\&A instruction datasets for training our model on the quality score regression task, referred to as \modelname{} (wo/multi-task). Additionally, we also leverage Q\&A instruction datasets for both quality score regression and quality classification tasks to train another version of the model, termed as \modelname{} (w/multi-task).


\begin{table*}[ht]
\centering
\small
\renewcommand\arraystretch{1.2}

\small
\vspace{-5pt}
\caption{In-sample performance of \modelname{} against competitive methods. All methods are trained on LSVQ$_\text{train}$ and evaluated on two intra-dataset (LSVQ$_\text{test}$ and LSVQ$_\text{1080p}$). W/wo refers to whether our model is trained on multi-tasks or not. PaQ-2-PiQ~\cite{Ying2020PaQ-2-PiQ} is an IQA model that pre-trained ResNet-18~\cite{he2016deep} on the LIVE-FB dataset~\cite{Ying2020PaQ-2-PiQ}. UNIQUE~\cite{zhang2021uncertainty} is an IQA model that pre-trained ResNet-50~\cite{he2016deep} on four fused IQA datasets~\cite{Fang2020spaq, koniq-10k, 2011bid, 2015livechallenge}. Our methods are marked in {\colorbox{baselinecolor}{gray}} and the best results are highlighted in \textbf{bold}.}
\setlength{\tabcolsep}{4.1mm}{\begin{tabular}{l:l:cc:cc}
\hline 
\multicolumn{1}{l}{\textit{Training Set:}  \textbf{LSVQ$_\text{train}$}} & \multicolumn{1}{c}{$  \to  $\textit{Testing Set:}} & \multicolumn{2}{c}{\textbf{ LSVQ\(_\text{test}\)}} & \multicolumn{2}{c}{\textbf{LSVQ\(_\text{1080p}\)}}\\ 
\hline
\textbf{Method}        &\multicolumn{1}{c:}{\bf Feature Extraction Modules} &SRCC &PLCC   &SRCC &PLCC  \\
\hline
TLVQM~\cite{Korhonen2019TLVQM} (TIP 2019)          &Handcrafted features &0.772&0.774  &0.589&0.616\\
VIDEVAL~\cite{tu2021VIDEVAL} (TIP 2021)             &Handcrafted features &0.794&0.783  &0.545&0.554\\
\hdashline
VSFA~\cite{VSFA} (ACMMM 2019)              &ResNet-50 based&0.801&0.796  &0.675&0.704\\
PVQ~\cite{lsvq} (CVPR 2021)                &PaQ-2-PiQ \& 3D ResNet-18 based  &0.827&0.828  &0.711&0.739\\
BVQA~\cite{li2022bvqa} (TCSVT 2022)              &UNIQUE \& SlowFast based   &0.852&0.854  &0.772&0.788\\
FAST-VQA~\cite{fast-vqa} (ECCV 2022)           &Video Swin-T   &0.876&0.877  &0.779&0.814\\
SimpleVQA~\cite{simple-vqa} (ACMMM 2022)         &ResNet-50 \& SlowFast based   &0.867&0.861  &0.764&0.803\\
MinimalisticVQA~\cite{sun2024analysis} (TPAMI, 2023)        &Swin-B \& SlowFast based   &0.881&0.879  &0.781&0.820\\ 
DOVER~\cite{wu2023dover} (ICCV 2023)    & ConvNeXt-T \& Video Swin-T based    &0.886&0.887  &0.795&0.830\\
\hdashline
Q-Align~\cite{q-align} (ICML, 2024)    &mPLUG-Owl-2 based   &0.883&0.882  &0.797&0.830\\
\rowcolor{mygray}
\textbf{\textsc{\modelname{}} (wo/multi-task)} &ViT \& SlowFast \& Llama-3 based    &0.913&0.914 & 0.879&0.898\\
\rowcolor{mygray}
\textbf{\textsc{\modelname{}} (w/multi-task)} &ViT \& SlowFast \& Llama-3 based &\textbf{0.916}&\textbf{0.919} & \textbf{0.891}&\textbf{0.899}\\
\hline

\end{tabular}}
\label{tab:insample}
\vspace{-5pt}
\end{table*}


\begin{table*}[ht]
\centering
\small
\renewcommand\arraystretch{1.18}

\caption{OOD performance of \modelname{} against competitive methods. All methods are trained on LSVQ$_\text{train}$ and evaluated on four OOD datasets including KoNViD-1k, YouTube-UGC, LIVE-VQC, and LIVE-YT-Gaming. \gray{NA} in Table refers to that the results cannot be produced.}

\setlength{\tabcolsep}{3.5mm}{\begin{tabular}{l:cc:cc:cc:cc}

\hline
\multicolumn{1}{c}{\textit{Training Set:}  \textbf{LSVQ$_\text{train}$}  $  \to  $\textit{Testing Set:}} & \multicolumn{2}{c}{\textbf{KoNViD-1k}} & \multicolumn{2}{c}{\textbf{YouTube-UGC}} & \multicolumn{2}{c}{\textbf{LIVE-VQC}}& \multicolumn{2}{c}{\textbf{LIVE-YT-Gaming}}\\ 
\hline
\multicolumn{1}{l:}{\textbf{Method}}        &\multicolumn{1}{c}{SRCC}  &\multicolumn{1}{c:}{PLCC}   &\multicolumn{1}{c}{SRCC} &\multicolumn{1}{c:}{PLCC}   &\multicolumn{1}{c}{SRCC}&\multicolumn{1}{c:}{PLCC}&\multicolumn{1}{c}{SRCC}&\multicolumn{1}{c}{PLCC}  \\
\hline 
TLVQM~\cite{Korhonen2019TLVQM} (TIP 2019)                 &0.732&0.724  &0.685&0.692  &0.670&0.691 &0.740&0.775  \\
VSFA~\cite{VSFA} (ACMMM 2019)                &0.784&0.794  &0.733&0.728  &0.753&0.795 &0.669&0.698  \\
VIDEVAL~\cite{tu2021VIDEVAL} (TIP 2021)&0.751&0.741  &0.687&0.709  &0.630&0.640 &\gray{NA}&\gray{NA}  \\
PVQ~\cite{lsvq} (CVPR 2021)                  &0.791&0.795  &0.742&0.754  &0.770&0.807 &\gray{NA}&\gray{NA}  \\
BVQA~\cite{li2022bvqa} (TCSVT 2022)                &0.839&0.830  &0.802&0.792  &0.816&0.824 &0.690&0.740  \\
SimpleVQA~\cite{simple-vqa} (ACM MM 2022)                &0.839& 0.841 &0.811&0.813  &0.730& 0.780&0.666& 0.734 \\
FAST-VQA~\cite{fast-vqa} (ECCV 2022)             &0.859&0.855  &0.730&0.747  &0.823&0.844 &0.620&0.666  \\
MinimalisticVQA~\cite{sun2024analysis} (TPAMI 2024)     &0.862&0.863  & 0.826&0.833  &0.798&0.837 &0.711&0.755  \\ 
Q-Align~\cite{q-align} (ICML, 2024)             &0.865&0.877 &0.834&0.848 &0.773&0.830 &0.611&0.687\\
\rowcolor{mygray}
\textbf{\textsc{\modelname{}} (wo/multi-task)} & 0.875&0.876 & \textbf{0.858}&\textbf{0.877} & \textbf{0.831}&\textbf{0.863} &0.783   &0.789 \\
\rowcolor{mygray}
\textbf{\textsc{\modelname{}} (w/multi-task)} & \textbf{0.901}&\textbf{0.902} & 0.748& 0.761 & 0.767&\ 0.805 &\textbf{0.816}   &\textbf{0.801} \\

\hline

\end{tabular}}
\label{tab:oodd}
\vspace{-10pt}
\end{table*}

\section{Experiment} \label{sec:experiment}


\subsection{Experimental Settings}
\subsubsection{Implementation details}
We utilize \emph{Llama-3-8b-Instruct}~\cite{dubey2024llama3} as the decoder, \emph{clip-vit-large-L/14}~\cite{radford2021learning} and \emph{SlowFast}~\cite{SlowFast} as the spatial and temporal encoders, respectively. We utilize paired Q\&A instruction prompts as our training data. During the training phase, the parameters of both the decoder and the vision encoders are frozen. We only train the spatial and temporal projectors to align modalities between visual and text tokens. Training experiments are conducted on eight NVIDIA A800 GPUs, while inference is performed on two NVIDIA A6000 GPUs. The batch size is set as $32$, with a learning rate of $0.001$, and training epochs are set as $6$ to achieve convergence. The video input resolution $H\times W$ is $224\times224$ and the patch size $p$ is typically set to $14$ for the ViT-L/14 backbone. The length of video chunk $\tau$ is configured to equal the video frame rate, \textit{i.e.}, one video chunk and one key frame are sampled per second.


\subsubsection{Evaluation criteria}
We select Spearman Rank-Order Correlation Coefficient (SRCC), and Pearson Linear Correlation Coefficient (PLCC) as metrics for performance evaluation, which indicate the prediction monotonicity and prediction accuracy. It should be noted that SRCC and PLCC range from 0 to 1.0, where larger values indicate better results.

\subsubsection{VQA datasets}
For instruction tuning and evaluation, we construct our Q\&A instruction prompts based on  five datasets: LSVQ~\cite{lsvq}, KoNViD-1k~\cite{konvid-1k}, YouTube-UGC~\cite{youtube-ugc}, LIVE-VQC~\cite{live-vqc}, LIVE-YT-Gaming~\cite{live-yt-gaming}, with details provided in Table~\ref{tab:dataset intro}. LSVQ, the largest VQA dataset in our study, comprises $38,811$ videos, covering a wide range of resolutions and frame rates. KoNViD-1K consists of $1,200$ videos of fixed resolution sourced from the Internet, primarily focusing on spatial distortions. LIVE-VQC includes $585$ videos captured directly by various mobile cameras, showcasing a broad range of natural scenes, different lighting conditions, and diverse motion levels. YouTube-UGC consists of $1,500$ YouTube video clips, each 20 seconds in duration, with resolutions spanning from 360p to 4K. Targeting streamed gaming videos, LIVE-YT-Gaming contains $600$ gaming videos belonging to $59$ games.

For the LSVQ dataset, we follow the original dataset splitting~\cite{lsvq}, in which the training set consists of $28,013$ videos, while $7,220$ and $3,561$ videos for the default test and test-1080p sets respectively. Taking the LSVQ dataset as the training set, we randomly sample 10\% videos from the training set for validation and report the results on the two intro test sets. For the remaining four smaller datasets, we use 5-fold cross-validation in fine-tuning experiments and report the results on the entire dataset for OOD evaluation.

\noindent
\subsubsection{Competing methods} 
We choose several representative VQA methods for comparison. The competing methods can be roughly classified into three groups: 1) three handcrafted models: TLVQM~\cite{Korhonen2019TLVQM}, VIDEVAL~\cite{tu2021VIDEVAL}, and RAPIQUE~\cite{tu2021rapique}; 2) seven DNN-based methods: VSFA~\cite{VSFA}, PVQ~\cite{lsvq}, BVQA~\cite{li2022bvqa}, FAST-VQA~\cite{fast-vqa}, SimpleVQA~\cite{simple-vqa}, MinimalisticVQA~\cite{sun2024analysis}, and DOVER~\cite{wu2023dover}; 3) an LMM-based method: Q-Align~\cite{q-align}. 

\begin{table*}[ht]
\centering
\small
\renewcommand\arraystretch{1.18}
\vspace{-5pt}
\caption{The fine-tuning performance of \modelname{} against competitive methods. All methods are trained on the LSVQ$_\text{train}$ and fine-tuned on four OOD datasets including KoNViD-1k, YouTube-UGC, LIVE-VQC, and LIVE-YT-Gaming. We performed a five-fold cross-validation to evaluate performance. The results presented in the table are the average values derived from these five experiments.}

\setlength{\tabcolsep}{10.3pt}{\begin{tabular}{l:cc:cc:cc:cc}

\hline
\multicolumn{1}{l}{\textit{Fine-tuning }\&\textit{ Testing Set:}} & \multicolumn{2}{c}{\textbf{KoNViD-1k}} & \multicolumn{2}{c}{\textbf{YouTube-UGC}} & \multicolumn{2}{c}{\textbf{LIVE-VQC}}& \multicolumn{2}{c}{\textbf{LIVE-YT-Gaming}}\\ 
\hline
\multicolumn{1}{l}{\textbf{Method}}        &\multicolumn{1}{c}{SRCC}  &\multicolumn{1}{c}{PLCC}   &\multicolumn{1}{c}{SRCC} &\multicolumn{1}{c}{PLCC}   &\multicolumn{1}{c}{SRCC}&\multicolumn{1}{c}{PLCC}&\multicolumn{1}{c}{SRCC}&\multicolumn{1}{c}{PLCC}  \\
\hline 

VSFA~\cite{VSFA} (ACMMM 2019)                &0.794 &0.799  &0.787&0.789  &0.718&0.771 &0.784 &0.819  \\
RAPIQUE~\cite{tu2021rapique} (OJSP 2021)  &0.811&0.819  &0.774&0.781  &0.740&0.764  &0.771&0.815    \\
BVQA~\cite{li2022bvqa} (TCSVT 2022)                &0.834&0.836  &0.818&0.826  &0.834&0.842 &\gray{NA}&\gray{NA}  \\
SimpleVQA~\cite{simple-vqa} (ACMMM 2022)            &0.856&0.860  &0.847&0.856  &0.845&0.859 &0.861&0.866  \\
FAST-VQA~\cite{fast-vqa} (ECCV 2022)                &0.891&0.892  &0.852&0.855  &0.849&0.865 &0.869&0.880    \\
DOVER~\cite{wu2023dover} (ICCV 2023)                &0.909&0.906  &0.890&0.891  &0.860&0.875 &0.852&0.868\\ 
MinimalisticVQA~\cite{sun2024analysis} (TPAMI 2024) &0.889&0.890  &0.890&0.891  &0.842&0.854 &0.857&0.888 \\
\rowcolor{mygray}
\textbf{\textsc{\modelname{}} (wo/multi-task)}               &\textbf{0.929}   &\textbf{0.930}     &\textbf{0.901}   &\textbf{0.897}    &\textbf{0.891}   &\textbf{0.903}    & \textbf{0.914}  &\textbf{0.901}  \\ 
\hline

\end{tabular}}
\label{tab:ft}
\vspace{-5pt}
\end{table*}

\begin{table*}[ht]
\centering
\small
\renewcommand\arraystretch{1.18}

\caption{The classification performance of \modelname{} against existing SOTA methods. All methods are trained on LSVQ$_\text{train}$ and evaluated on two in-sample datasets (LSVQ$_\text{test}$ and LSVQ$_\text{1080p}$) and four OOD datasets (KoNViD-1k, YouTube-UGC, LIVE-VQC, and LIVE-YT-Gaming).}
\centering
\setlength{\tabcolsep}{4.9pt}{\begin{tabular}{l:cccc:cccc:cccc}

\hline
\multicolumn{1}{c}{\textit{Training Set:}  \textbf{LSVQ$_\text{train}$}  $  \to  $\textit{Testing Set:}} & \multicolumn{4}{c}{\textbf{ LSVQ\(_\text{test}\)}}& \multicolumn{4}{c}{\textbf{ LSVQ\(_\text{1080p}\)}}& \multicolumn{4}{c}{\textbf{KoNViD-1k}} \\
\hline
\multicolumn{1}{l:}{\textbf{Level}}        &\multicolumn{1}{c}{Poor}  &\multicolumn{1}{c}{Fair}   &\multicolumn{1}{c}{Good} &\multicolumn{1}{c:}{Total}   &\multicolumn{1}{c}{Poor}  &\multicolumn{1}{c}{Fair}   &\multicolumn{1}{c}{Good} &\multicolumn{1}{c:}{Total}   &\multicolumn{1}{c}{Poor}  &\multicolumn{1}{c}{Fair}   &\multicolumn{1}{c}{Good} &\multicolumn{1}{c}{Total}   \\
\hline 
Q-Align~\cite{q-align} (ICML, 2024)                 &0.880&0.288&\textbf{0.931}&0.702    &0.558&0.142&0.973&0.561    &0.922&0.415&0.797&0.709 \\ 
MinimalisticVQA~\cite{sun2024analysis} (TPAMI 2024) &0.797&0.677&0.756&0.744&	0.711&0.668&0.614&0.664&	0.841&0.716&0.248&0.602 \\
\rowcolor{mygray}
\textbf{\textsc{\modelname{}} (w/multi-task)}      &\textbf{0.914}&\textbf{0.843}&0.925&\textbf{0.950}    &\textbf{0.953}&\textbf{0.897}&\textbf{0.977}&\textbf{0.955}    &\textbf{0.940}&\textbf{0.943}&\textbf{0.967}&\textbf{0.950}    \\ 
\hline
\hline
\multicolumn{1}{r}{\textit{Testing Set:}} & \multicolumn{4}{c}{\textbf{YouTube-UGC}}& \multicolumn{4}{c}{\textbf{LIVE-VQC}}& \multicolumn{4}{c}{\textbf{LIVE-YT-Gaming}} \\
\hline
\multicolumn{1}{l:}{\textbf{Level}}        &\multicolumn{1}{c}{Poor}  &\multicolumn{1}{c}{Fair}   &\multicolumn{1}{c}{Good} &\multicolumn{1}{c:}{Total}   &\multicolumn{1}{c}{Poor}  &\multicolumn{1}{c}{Fair}   &\multicolumn{1}{c}{Good} &\multicolumn{1}{c:}{Total}   &\multicolumn{1}{c}{Poor}  &\multicolumn{1}{c}{Fair}   &\multicolumn{1}{c}{Good} &\multicolumn{1}{c}{Total}   \\
\hline 

Q-Align~\cite{q-align} (ICML, 2024)                 &\textbf{0.976}&0.294&0.034&0.431	&0.451&0.788&0.467&0.568	&0.389&\textbf{0.914}&0.029&0.440    \\

MinimalisticVQA~\cite{sun2024analysis} (TPAMI 2024) &0.923&0.395&0.544&0.620	&0.653&0.347&\textbf{0.915}&0.641	&0.390&0.626&0.549&0.522    \\

\rowcolor{mygray}
\textbf{\textsc{\modelname{}} (w/multi-task)}      &0.960&\textbf{0.830}&\textbf{0.727}&\textbf{0.838}    &\textbf{0.933}&\textbf{0.839}&0.854&\textbf{0.875}    &\textbf{0.667}&0.514&\textbf{0.684}&\textbf{0.625}    \\ 
\hline
\end{tabular}}
\label{tab:class-ood}
\vspace{-8pt}
\end{table*}

\subsection{Performance Comparison}
\noindent
\subsubsection{Quality scoring task}
To comprehensively evaluate our proposed method, we assess our model with an in-sample, OOD, and fine-tuning manner. 

Table~\ref{tab:insample} presents the in-sample results for BVQA models on the LSVQ and LSVQ-1080p datasets. All models are trained on the LSVQ training set (LSVQ\(_\text{train}\)) and evaluated on LSVQ\(_\text{test}\) and LSVQ\(_\text{1080p}\) test sets. It is observed that \modelname{} consistently outperforms all competing methods in terms of both SRCC and PLCC. Specifically, for LSVQ\(_\text{test}\), \modelname{} (wo/multi-task) achieves an SRCC of $0.913$ and a PLCC of $0.914$, outperforming SOTA models such as DOVER~\cite{wu2023dover} and Q-Align~\cite{q-align}. Further, \modelname{} (w/multi-task) performs better than \modelname{} (wo/multi-task), which indicates that multi-task learning for quality score regression and quality level classification enhances the quality understanding of videos for LMMs, then improves the prediction performance of quality scores. On the LSVQ\(_\text{1080p}\) set, \modelname{} also leads with an SRCC of $0.879$ and a PLCC of $0.898$, significantly surpassing DNN-based models like DOVER~\cite{wu2023dover}, and LMM-based method like Q-Align~\cite{q-align}. These results show that our proposed \modelname{} achieves the best performance in both SRCC and PLCC across different test sets, demonstrating its effectiveness in VQA tasks. 

We conduct OOD evaluations to further evaluate the generalization capability of our \modelname{}. Table~\ref{tab:oodd} shows that \modelname{} (wo/multi-task) significantly outperforms the other models on KoNViD-1K~\cite{konvid-1k}, YouTube-UGC~\cite{youtube-ugc}, LIVE-VQC~\cite{youtube-ugc} and LIVE-YT-Gaming~\cite{live-yt-gaming} datasets. For KoNViD-1k, \modelname{} (wo/multi-task) achieves an SRCC of $0.875$ and a PLCC of $0.876$, outperforming other LMM-based models such as Q-Align ($0.865$ and $0.877$), along with MinimalisticVQA ($0.862$ and $0.863$), the best DNN-based models. For YouTube-UGC and LIVE-VQC datasets, which present significant challenges due to poor exposure conditions and a variety of motion types, our proposed \modelname{} (wo/multi-task) achieves an SRCC/PLCC of $0.858/0.877$ on the YouTube-UGC dataset and $0.831/0.863$ on the LIVE-VQC dataset. This superior performance can largely be attributed to the generalization capabilities of LLMs, which are pre-trained on the vast unlabeled text corpus. This extensive pre-training enables \modelname{} to effectively understand the nuances of MOS, adapt to diverse prompts in VQA, and generalize across various datasets.
For the LIVE-YT-Gaming dataset that focuses on a specific type of video, \modelname{} (wo/multi-task) has also achieved SOTA performance, demonstrating its excellent generalization capability. However, \modelname{} (w/multi-task) exhibits substantial performance fluctuations. It achieves exceptionally excellent results on the KoNViD-1K and LIVE-YT-Gaming datasets but falls short of SOTA performance on the YouTube-UGC and LIVE-VQC datasets. 
The reason is that the quality distribution of videos differs between the training set LSVQ and two test sets, YouTube-UGC and LIVE-VQC are different, thus making the quality levels of the same video vary across these sets. So, multi-task training with quality level classification may lead to LMM-VQA misunderstanding the quality levels of videos in YouTube-UGC and LIVE-VQC, thereby affecting their quality scores.

Table~\ref{tab:ft} illustrates the fine-tuning performance against different methods. From the results, we can see that \modelname{} (wo/multi-task) maintains SOTA performance across all benchmark datasets. For the LIVE-VQC dataset, our \modelname{} yields an SRCC and PLCC of $0.891$ and $0.903$, demonstrating improvements of 7.2\% and 4.6\% compared with the model without fine-tuning, respectively. For the LIVE-YT-Gaming dataset, \modelname{} still attains an SRCC and PLCC of $0.914$ and $0.901$, showing improvements of 12\%, 12.5\% compared to the model without fine-tuning. Further, compared to the previous SOTA method, MinimalisticVQA, we obtain an average improvement of 4.6\% SRCC and 3.1\% PLCC in performance. These results show that LMM-VQA can be successfully trained on small-scale VQA datasets without overfitting.

\noindent
\subsubsection{Quality classification task}
As we introduce quality level classification as an auxiliary task for multi-task training, we evaluate the performance of the proposed model in predicting quality levels. We calculate the prediction accuracy across four scenarios: 'Poor`, 'Fair`, 'Good`, and 'Total`, where 'Total` refers to accuracy calculated across all test quality levels. All models are trained on the same dataset (LSVQ$_\text{train}$) and evaluated on datasets including LSVQ\(_\text{test}\), LSVQ\(_\text{1080p}\), KoNViD-1k, YouTube-UGC, LIVE-VQC, and LIVE-YT-Gaming datasets. Table~\ref{tab:class-ood} shows that \modelname{} significantly outperforms the traditional DNN-based model like MinimalisticVQA and LMM-based model like Q-Align across several datasets.

\subsection{Ablation Study}

\subsubsection{Temporal tokens}
Researchers~\cite{Korhonen2019TLVQM, li2022bvqa,sun2024analysis} have demonstrated that the temporal relationship between frames of the video is very important for VQA. We model the temporal relationship to enhance the LMM's capabilities of capturing the temporal distortion cues in videos. Thus, to demonstrate the rationality of \modelname{}, we discuss the effect of temporal tokens in our ablation study.
Table~\ref{tab:variant} (I) and (II) compare the performance of our proposed \modelname{} on two conditions: with temporal tokens (I) and without temporal tokens (II). We compare their performance on both the LSVQ test and test-1080p sets. From the results, we observe that the incorporation of temporal tokens consistently enhances the performance metrics across both datasets. For LSVQ\(_{\text{test}}\), the metrics PLCC/SRCC rise from $0.855/0.887$ to $0.889/0.891$, respectively. Moreover, by integrating temporal tokens into Llama-3, we achieve improvements of $2.3\%$ (PLCC) and $3.0\%$ (SRCC) in the prediction accuracy of LSVQ\(_{\text{1080p}}\).

\begin{figure*}[ht]
    \vspace{-10pt}
    \centering
    \includegraphics[width=0.9\linewidth]{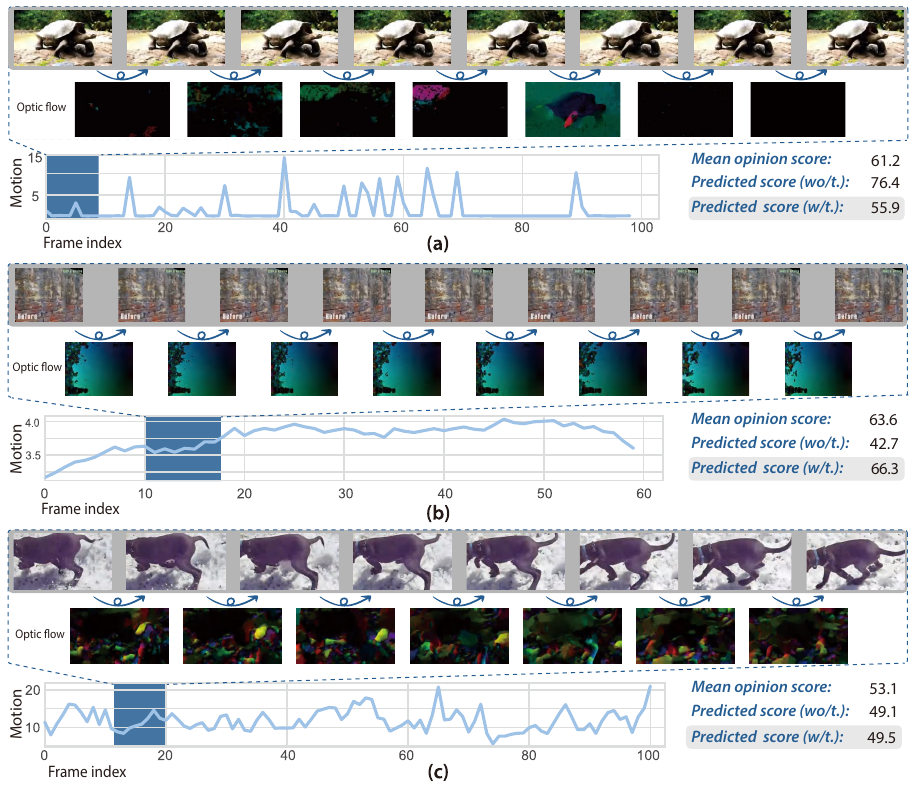}
    \captionsetup{size=small}
    \caption{\textbf{Visualization of predictions of the proposed model on LSVQ datasets}. We discuss three scenarios to discuss the effect of temporal tokens for VQA. (a) For lagging videos, introducing temporal tokens avoids the overestimation of quality scores.; (b) For complex textures of videos, motion tokens prevent underestimation of quality scores. (c) In some smooth videos, adding motion tokens shows no improvement in prediction accuracy.}
    \label{fig:motion case study}
    \vspace{-10pt}
\end{figure*}

We also visualize several predictions of \modelname{} in Fig.~\ref{fig:motion case study}. We use TVL1 algorithm~\cite{TV-L1} to compute optical flow values. These values are subsequently scalarized and spatially averaged to derive motion values between frames. 
Specifically, in case (a), the video exhibits stutter, a condition that has been discussed above and poses difficulties for several existing methods such as~\cite{q-align}. Under this condition, the model without temporal tokens significantly overestimates the scores of videos. By introducing temporal tokens, the accuracy of \modelname{} is substantially improved. In case (b), for videos containing complex textures, the model that lacks temporal tokens is significantly lower than the labeled MOS. This discrepancy arises because, in the absence of temporal information, video frames with limited resolution may cause the model to mistake complex textures for noise. Such misinterpretations commonly occur in scenes featuring elements like rainy weather, grasslands, and textured wall surfaces. By introducing temporal tokens, the model significantly improves its prediction accuracy. For case (c), it is a normal smooth video, and the results indicate that adding temporal information does not improve the accuracy of predictions.

\begin{table}[ht]
\setlength\tabcolsep{4pt}
\renewcommand\arraystretch{1.2}
\small
\vspace{-5pt}
\caption{Ablation study on motion tokens and model structure of proposed methods. Both SRCC and PLCC metrics are calculated on LSVQ test sets. All micro-designs further improve the performance of the proposed method.  }

\centering
\resizebox{\linewidth}{!}{
    \begin{tabular}{lcccc}
    \hline
    \multicolumn{1}{c}{\multirow{2}{*}{\textbf{Micro-design}}} & \multicolumn{2}{c}{\textbf{LSVQ$_\text{test}$}} & 
    \multicolumn{2}{c}{\textbf{LSVQ$_\text{1080p}$}}\\
    \cline{2-5} 
       &SRCC   &PLCC   &SRCC   &PLCC          \\
    \hline
    (\uppercase\expandafter{\romannumeral1}): Llama-2 as backbone, ViT-L/14 &0.855  &0.887  &0.833  &0.864           \\ 
    (\uppercase\expandafter{\romannumeral2}): (\uppercase\expandafter{\romannumeral1}) + temporal tokens     &0.889  &0.891  &0.858  &0.884           \\
    \gray{Improvement over (\uppercase\expandafter{\romannumeral1})} & \gray{+3.9\%} & \gray{+0.4\%} &\gray{+3.0\%}& \gray{+2.3\%} \\
    (\uppercase\expandafter{\romannumeral3}): (\uppercase\expandafter{\romannumeral2}) + Llama-2 $\to$ Llama-3  &0.913 & 0.914 & 0.879  & 0.898           \\
    \gray{Improvement over (\uppercase\expandafter{\romannumeral1})} & \gray{+6.7\%} &\gray{+3.0\%} & \gray{+5.5\%} & \gray{+3.9\%} \\
    \gray{Improvement over (\uppercase\expandafter{\romannumeral2})} & \gray{+2.6\%} &\gray{+2.6\%} &\gray{+2.4\%} & \gray{+1.6\%} \\
    \rowcolor{mygray}
    (\uppercase\expandafter{\romannumeral4}): (\uppercase\expandafter{\romannumeral3}) + multi-task   &\cellcolor{mygray} 0.916  &\cellcolor{mygray} 0.919 &\cellcolor{mygray} 0.891 &\cellcolor{mygray} 0.899 \\
    \gray{Improvement over (\uppercase\expandafter{\romannumeral1})} & \gray{+7.1\%} &\gray{+3.6\%} &\gray{+7.0\%} & \gray{+4.1\%} \\
    \gray{Improvement over (\uppercase\expandafter{\romannumeral3})} & \gray{+0.3\%} &\gray{+0.5\%} &\gray{+1.4\%} & \gray{+0.1\%} \\
    \hline
    
    \end{tabular}
}
\label{tab:variant}
\vspace{-10pt}
\end{table}

\subsubsection{The length of temporal tokens}
Next, we conduct experiments with the length of temporal tokens for further investigation. With the temporal features of SlowFast~\cite{SlowFast}, we set up $N_t=256$ as the token length of the slow feature. We select the default test set of LSVQ for verification. Table~\ref{tab:ablation_2} lists several token lengths of the slow feature. The results clearly show that performance consistently improves across SRCC/PLCC metrics as $N_t$ increases. Specifically, when $N_t$ is increased to $64$, we observe an obvious improvement of approximately $3\%$ compared to $N_t=4$. The model achieves optimal performance in SRCC and PLCC metrics with $N_t=256$ per frame of video. However, this configuration results in an increased running time of $3.99$ seconds per video, which is impractical. Thus, we select $N_t=64$ as a trade-off, identifying it as the optimal number of temporal tokens.

\subsubsection{Decoders} 
Next, we will discuss the effect of decoders. We select Llama-2-7b and Llama-3-7b for comparison. Table~\ref{tab:variant} (II) and (III) present the performance of two foundational models on two datasets: LSVQ\(_\text{test}\) and LSVQ\(_\text{1080p}\). The results indicate that the \modelname{} with Llama-3 outperforms the Llama-2-based model across both datasets. Specifically, the performance metrics (PLCC/SRCC) for Llama-3 on LSVQ\(_{\text{test}}\) increase from $0.889/0.891$ to $0.913/0.914$, and on the LSVQ\(_{\text{1080p}}\) from $0.858/0.884$ to $0.879/0.898$.  

\subsubsection{The task of quality classification}
Table~\ref{tab:variant} (III) and (IV) compare the performance of our proposed \modelname{} on two conditions: with and without multi-task training strategies respectively. The results illustrate that introducing classification tasks for training gains $7.1\%$ SRCC and $3.6\%$ benefits compared to baseline methods, suggesting that the multi-task training strategy can significantly improve model performance.

\begin{table*}[ht]
    \centering
    \vspace{-5pt}
    \caption{\textbf{Ablation study of proposed methods on LSVQ test sets}. (a) Ablation study of the spatial projector. We select MLP, Perceiver, and ViT modules for comparison; (b) Ablation study on training data proportions. We design four scenarios with varying fractions of training data for evaluation. For example, we randomly allocate a 25\% subset of training datasets, and test on LSVQ default test set; (c) Ablation study on the length of temporal tokens, test on LSVQ default test set. RT refers to the running time on a specific video. Default settings are marked in {\colorbox{baselinecolor}{gray}}.}
    \vspace{-5pt}
    \label{tab:ablation_2}
    \subfloat[Spatial projector.
    \label{tab:spatial projector}
    ]{
    \centering
    \begin{minipage}{0.35\linewidth}{\begin{center}
    \tablestyle{4pt}{1.3}
    \begin{tabular}{y{30}x{20}x{20}x{20}x{20}}
    \hline
    &\multicolumn{2}{c}{LSVQ$_\text{test}$}& \multicolumn{2}{c}{LSVQ$_\text{1080p}$}\\
    &SRCC&PLCC &SRCC&PLCC \\
    \hline
    MLP                     &0.904&0.904  &0.866&0.892           \\ 
    Perceiver               &0.892&0.891  &0.844&0.878         \\
    \baseline{ViT}                     &\baseline{0.913} &\baseline{0.914} &\baseline{0.879}  &\baseline{0.898}        \\
    \hline
    \end{tabular}
    \end{center}
    }\end{minipage}
    }
    \subfloat[Proportions of training data.
    \label{tab:training data}
    ]{
    \centering
    \begin{minipage}{0.28\linewidth}{\begin{center}
    \tablestyle{4pt}{1.3}
    \begin{tabular}{y{30}x{30}x{30}}
    \hline
    &\multicolumn{1}{c}{SRCC}& \multicolumn{1}{c}{PLCC}\\
    \hline
    25\%      &0.758  &0.801
            \\
    50\%      &0.856  &0.876            \\
    75\%      &0.878   &0.881         \\
    \baseline{100\%}    &\baseline{0.913}  &\baseline{0.914 } \\
    \hline
    \end{tabular}
    \end{center}
    }\end{minipage}
    } 
    \subfloat[The length of temporal tokens.
    \label{tab:temporal token}
    ]{
    \centering
    \begin{minipage}{0.32\linewidth}{\begin{center}
    \tablestyle{4pt}{1.3}
    \begin{tabular}{y{30}x{30}x{30}x{30}}
    \hline
    &\multicolumn{1}{c}{SRCC}& \multicolumn{1}{c}{PLCC}& \multicolumn{1}{c}{RT/s}\\
    \hline
    4       &0.879&0.879  &       1.98   \\
    16      &0.886&0.887  &       2.89   \\
    \baseline{64}      &\baseline{0.904}& \baseline{0.904}  &   \baseline{3.26}  \\
    256     &0.913& 0.914 & 3.99\\
    \hline
    \end{tabular}
    \end{center}
    }\end{minipage}
    }  
\end{table*}

\subsubsection{Proportions of training data}
Recent work~\cite{kaplan2020scaling} has found power-law scalings relationships between performance and dataset size in LLMs. In this study, we examine the performance of our model across various proportions of training data to determine whether the current amount of training data is sufficient. Table~\ref{tab:ablation_2} illustrates the performance of \modelname{} across different proportions of training data. The results show a clear increasing trend indicating that increasing the proportion of training data generally enhances the model's performance for VQA. Specifically, performance steadily improves from $0.758$ at $25\%$ of training data to $0.913$ at $100\%$ of training data. We can infer that the current volume of training data is insufficient. Collecting additional data for training may yield greater benefits from LMM-based VQA methods.

\subsubsection{Modality Projection} 
Aligning representations from external modalities into language space remains a challenging task. In our proposed \modelname{}, we integrate spatial and temporal tokens into the language space using two projectors, as shown in Figure~\ref{fig:overall}. We use a ViT-based model for the spatial projector and an MLP layer for the temporal projector. Here, we discuss the impact of different spatial projectors, keeping the temporal projector unchanged. We compare three typical methods for modality projection, including Perceiver~\cite{jaegle2021perceiver}, MLP, and our proposed ViT-based module. As depicted in Table~\ref{tab:ablation_2}, our proposed ViT module achieves significant gains. Compared with Perceiver~\cite{jaegle2021perceiver}, our proposed ViT projector can be further enhanced and attains peak performance in all predicted metrics. For LSVQ\(_{\text{1080p}}\), our proposed ViT projector stands out with SRCC/PLCC of $0.879/0.898$, significantly surpassing MLP ($0.866/0.892$) and Perceiver ($0.844/0.878$). These findings suggest that the ViT projection layer is more effective in capturing and representing the relevant features, leading to superior performance. 

\begin{table*}[ht]
\renewcommand\arraystretch{1.2}
\small
\centering
\caption{Quantitative results on VCGbench.DEFAULT SETTINGS ARE MARKED IN GRAY .}
\setlength{\tabcolsep}{3.6mm}{
\begin{tabular}{lcccccc}
    \hline
    \multirow{2}{*}{\textbf{Method}} & \textbf{Using} &  \textbf{Information} & \textbf{Detailed} & \textbf{Contextual} & \textbf{Temporal} & \multirow{2}{*}{\textbf{Consistency}}\\
    &   \textbf{Subtitles} &  \textbf{Correctness} & \textbf{Orientation} & \textbf{Understanding} & \textbf{Understanding} & \\
    \hline
    LLaMA Adapter~\cite{llama-adapter}      & \xmark &  2.03 & 2.32 & 2.30 & 1.98 & 2.15 \\
    Video LLaMA~\cite{video-llama}          & \xmark &  1.96 & 2.18 & 2.16 & 1.82 & 1.79 \\
    Video Chat ~\cite{li2024videochat}      & \xmark &  2.23 & 2.50 & 2.53 & 1.94 & 2.24 \\
    Video-ChatGPT~\cite{maaz2023video}      & \xmark &  2.40 & 2.52 & 2.62 & 1.98 & 2.37 \\
    BT-Adapter-7B ~\cite{liu2023all}        & \xmark & 2.68  & 2.69 & 3.27 & 2.34 & 2.46 \\ 
    LLaMA-VID-7B~\cite{li2023llama}         & \xmark & 2.96  & 3.00 & \textbf{3.53} & 2.46 & 2.51 \\
    \rowcolor{mygray}
    \textbf{\modelname{}}                           & \xmark & \textbf{3.03}  & \textbf{3.07}  & \textbf{3.53} & \textbf{2.83} &\textbf{2.53} \\
  \hline
\end{tabular}
}
\vspace{-10pt}
\label{tab:benchmark_videochatgpt}
\end{table*}


\subsection{Performance evaluation on general benchmark}
To verify that our proposed model can also benefit general video understanding tasks, we assess its performance on the general VCGBench~\cite{maaz2023video}, detailed in Table~\ref{tab:benchmark_videochatgpt}. 

\subsubsection{Training pipeline}
For the general benchmark, we propose a special version of \modelname{}, which is trained with two stages, \textit{i.e.}, modality alignment, and instruction tuning. For modality alignment, we train a spatial projector and a temporal projector, which projects the spatial and temporal features to the LLM's language space. We leverage a combined image captioning dataset that includes images from LAION~\cite{schuhmann2021laion}, Conceptual Captions~\cite{sharma2018conceptual}, and SBU~\cite{ordonez2011im2text} to align the visual feature with LLM’s input space. In this stage, we primarily optimize the projectors in Fig.~\ref{fig:overall}, while freezing the pre-trained modules like the LLM and vision encoders. In the stage of instruction tuning, we construct pair-wised instruction datasets to enhance the multimodal understanding of LLMs. In particular, we take multiple frames as input and design the predefined prompts in the following template:
\texttt{Human: <prompt><image-1><image-2>..<image-N>, Assistant: <answer>}. The detailed contents of \texttt{<prompt>} and \texttt{<answer>} can be found in ~\cite{maaz2023video}. Further, we represent each image with 64 spatial tokens and 1 temporal token for Llama-3. In instruction tuning, all the modules are optimized except the frozen vision encoder. 
\subsubsection{Results analysis}
The VCGBench encompasses five critical aspects: information correctness, detail orientation, contextual understanding, temporal understanding, and consistency. From the results, we can see our model achieves SOTA performance in temporal understanding dimensions, which verified that our model can utilize the temporal tokens to improve video understanding. Further, the reason why our model can facilitate video understanding is that we encode both spatial and temporal features in SlowFast~\cite{SlowFast} into temporal tokens, so that those tokens contain the high-level temporal representation of videos. Then, these tokens introduce additional all-frame visual representations that are not captured by vision encoders with fixed frames. In the other evaluation dimensions, our model is comparable with the previous methods without subtitles.

\section{Conclusion and Future Discussions}

In this paper, we present \modelname{}, an LMM-based VQA model that aims at better modeling spatial and temporal relationships to predict accurate scores and levels for VQA. \modelname{} takes video frames as input, and generates the responses of quality scores with the quality levels. The process initiates with two vision encoders that separately transform input frames into spatial and temporal features. To better capture temporal distortions and align different modalities, we design two spatial and temporal projectors to integrate visual tokens into the language space for modality alignment. We utilize Llama-3 as our language decoder, the aligned visual tokens and quality prompt tokens are aggregated as input for Llama-3 to generate the final answers. The proposed \modelname{} has reached SOTA performance on five popular VQA datasets and achieved excellent generalization ability among them.

However, unintended risks may still exist when \modelname{} is unable to adequately analyze observations due to a limited understanding of quality regression, potentially leading to invalid outcomes in the VQA task. To address this issue, we may need to focus on improving the quality and diversity of the multimodal training data and refining the LMM’s understanding of the relationship between video and quality score concepts. On the other hand, we can incorporate human feedback into \modelname{} so that it can optimize the prediction performance of the quality scores iteratively. In the future, we will also discuss the reasoning process of \modelname{}'s generation to solve more complex problems including the better coverage of non-natural contents, and dealing with ambiguous quality scores.

\noindent

{\small
	\bibliographystyle{ieee_fullname}
	\bibliography{egbib}
}
\begin{IEEEbiography}[{\includegraphics[width=1in,height=1.25in,clip,keepaspectratio]{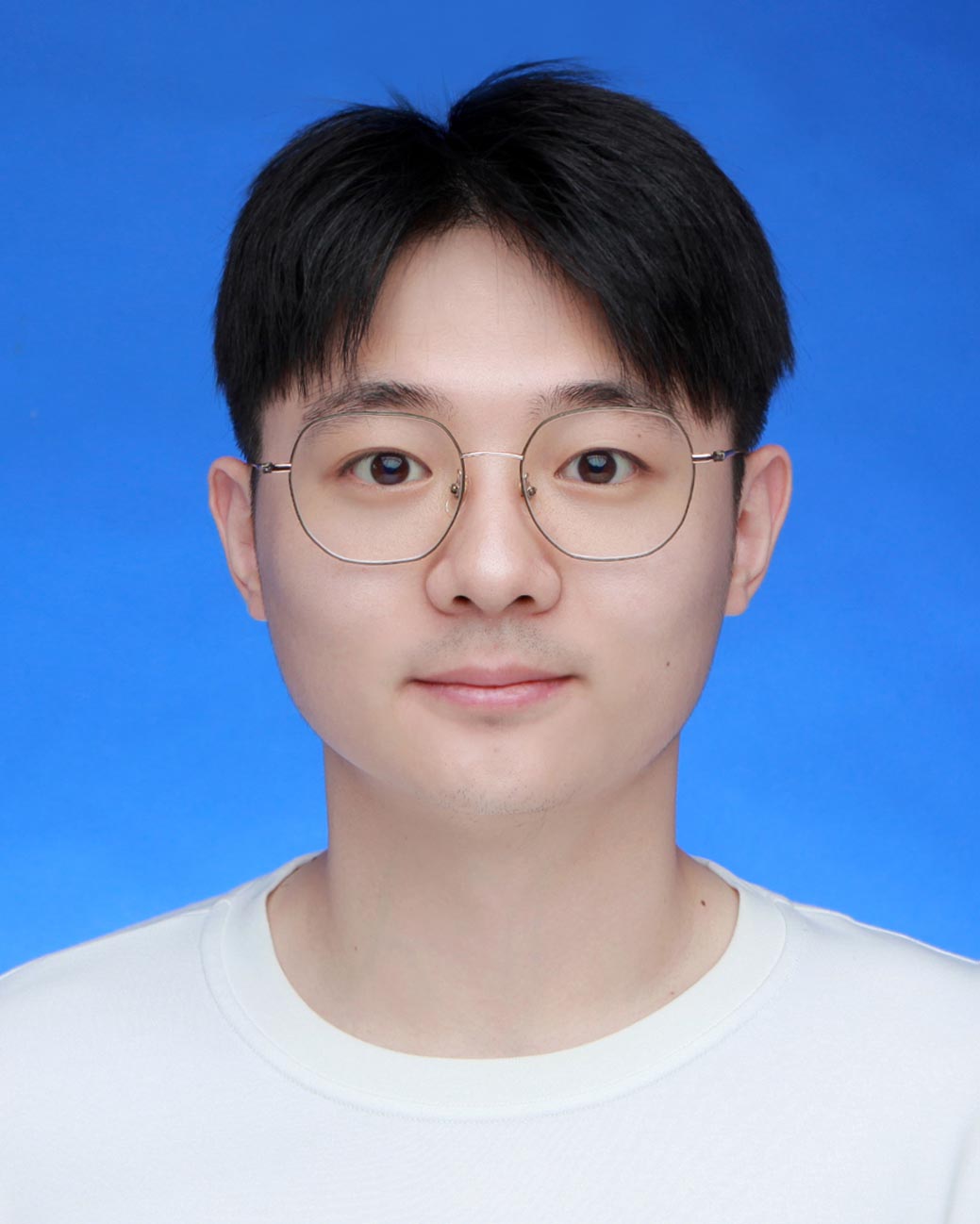}}]{Qihang Ge}
received the B.E. degree from Huazhong University of Science and Technology, Wuhan, China, in 2019. He is currently pursuing a Ph.D. degree with the Institute of Image Communication and Network Engineering, Shanghai Jiao Tong University. His research interests include video quality assessment and multimodal signal processing.
\vspace{-30pt}
\end{IEEEbiography}

\begin{IEEEbiography}[{\includegraphics[width=1in,height=1.25in,clip,keepaspectratio]{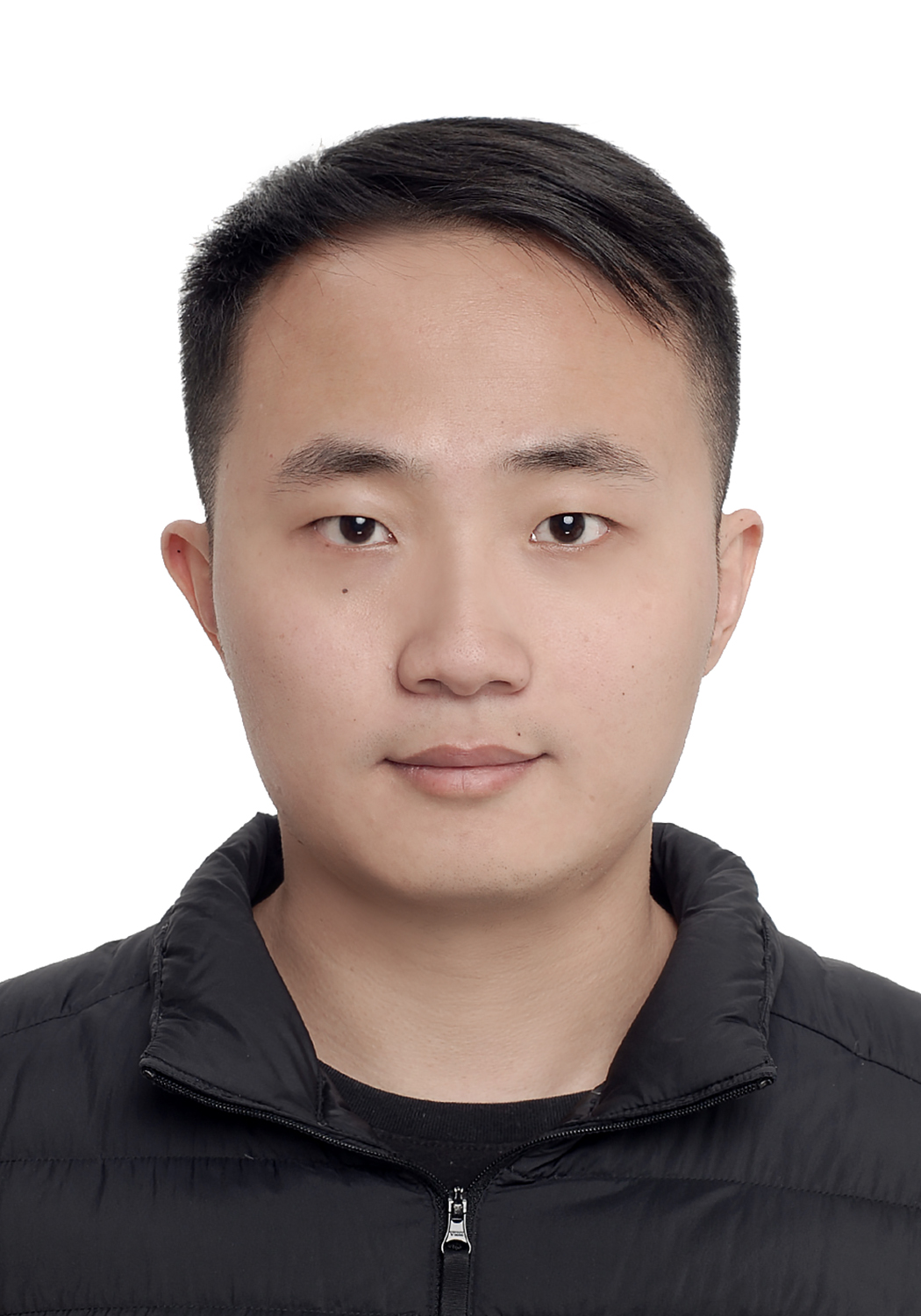}}]{Wei Sun}
received the B.E. degree from the East China University of Science and Technology, Shanghai, China, in 2016, and the Ph.D. degree from Shanghai Jiao Tong University, Shanghai, China, in 2023. He is currently a Post-Doctoral Fellow with Shanghai Jiao Tong University. His research interests include image quality assessment, perceptual signal processing, and mobile video processing.
\vspace{-30pt}
\end{IEEEbiography}

\begin{IEEEbiography}[{\includegraphics[width=1in,height=1.25in,clip,keepaspectratio]{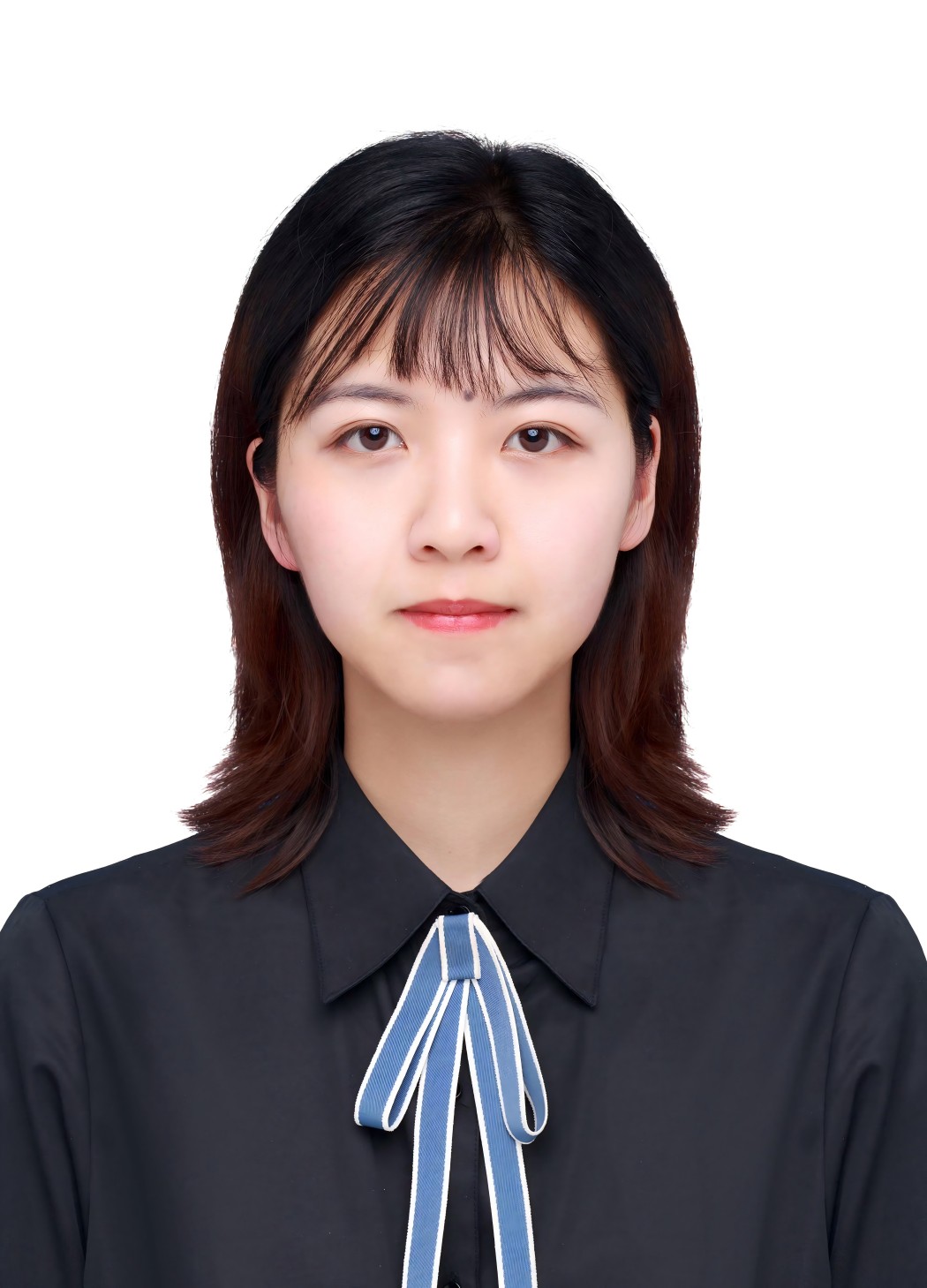}}]{Yu Zhang}
received the M.E. degree from Southeast University, Nanjing, China, in 2022. She is currently pursuing a Ph.D. degree with the MoE Key Laboratory of Artificial Intelligence, AI Institute, Shanghai Jiao Tong University. Her research interests include AI for science and representation learning.
\vspace{-30pt}
\end{IEEEbiography}

\begin{IEEEbiography}[{\includegraphics[width=1in,height=1.25in,clip,keepaspectratio]{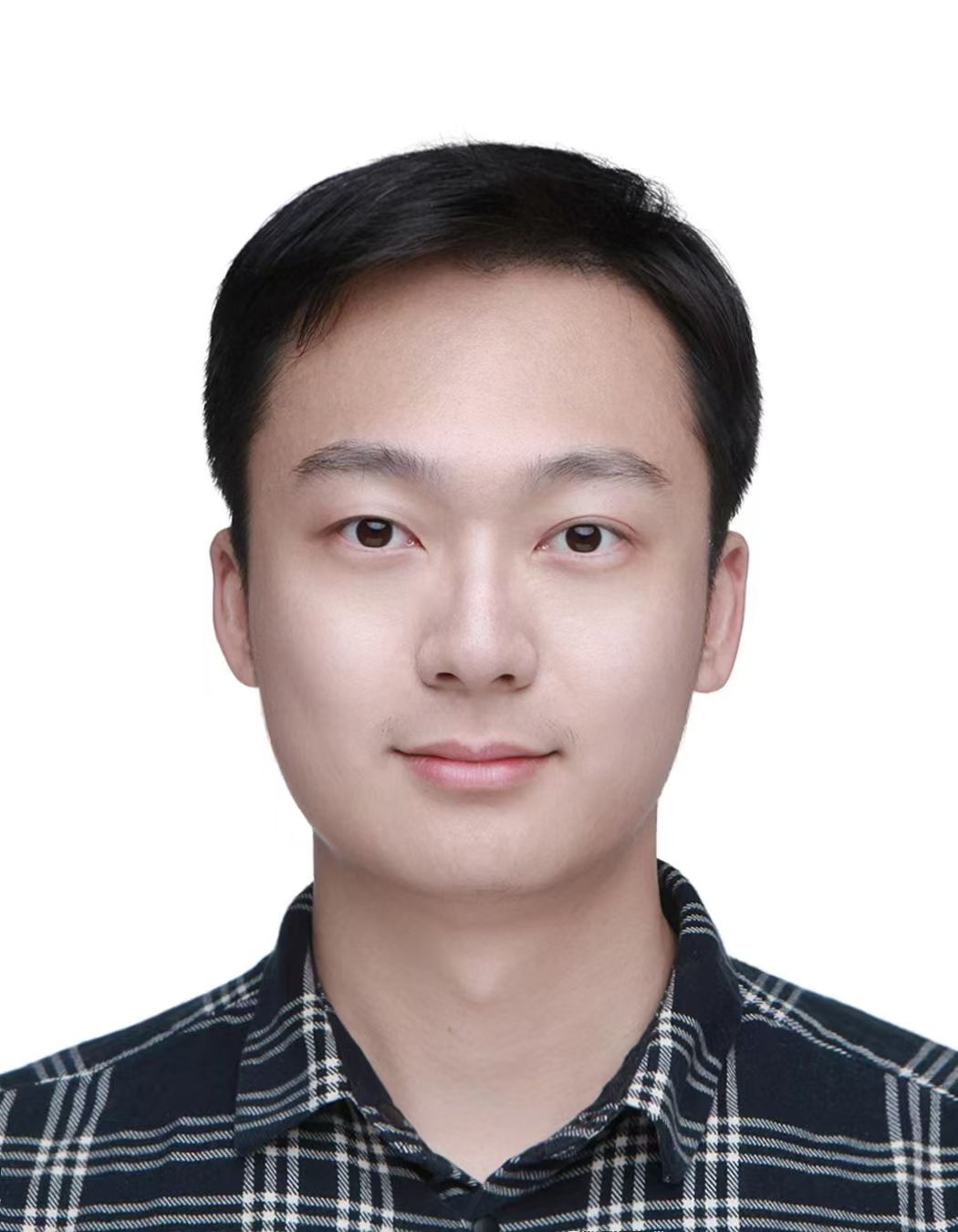}}]{Yunhao Li}
received the BE degree from Beihang University, Beijing, China, in 2019. He is currently working toward a PhD degree at the Institute of Image Communication and Network Engineering, Shanghai Jiao Tong University, Shanghai, China. His research interests include human-centric understanding and generation.
\vspace{-30pt}
\end{IEEEbiography}

\begin{IEEEbiography}[{\includegraphics[width=1in,height=1.25in,clip,keepaspectratio]{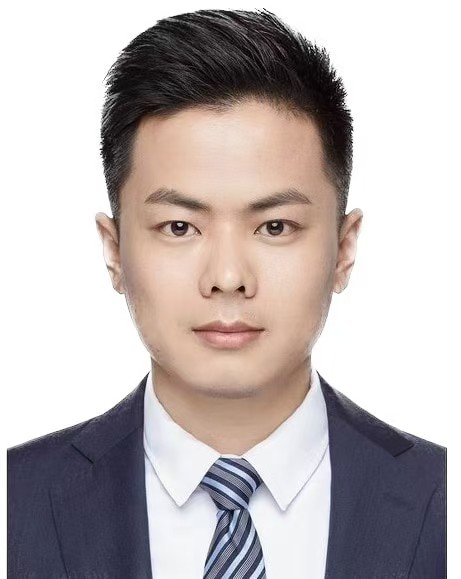}}]{Zhongpeng Ji}
received the Ph.D. degree from Shanghai Institute of Technical Physics, Chinese Academy of Sciences, Shanghai, China. He is currently working for Huawei Kirin Chipset Solution, Shenzhen, China.
\vspace{-30pt}
\end{IEEEbiography}

\begin{IEEEbiography}[{\includegraphics[width=1in,height=1.25in,clip,keepaspectratio]{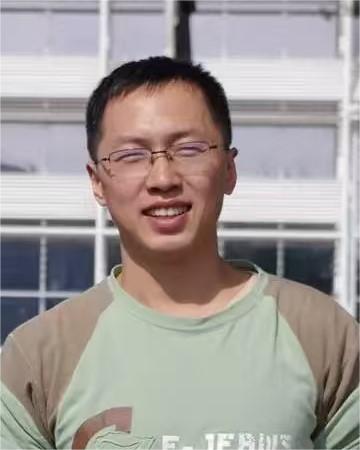}}]{Fengyu Sun}
received the B.E. and M.E. degrees from Tongji University, Shanghai, China. He is currently working for Huawei Kirin Chipset Solution, Shenzhen, China.
\vspace{-30pt}
\end{IEEEbiography}

\begin{IEEEbiography}[{\includegraphics[width=1in,height=1.25in,clip,keepaspectratio]{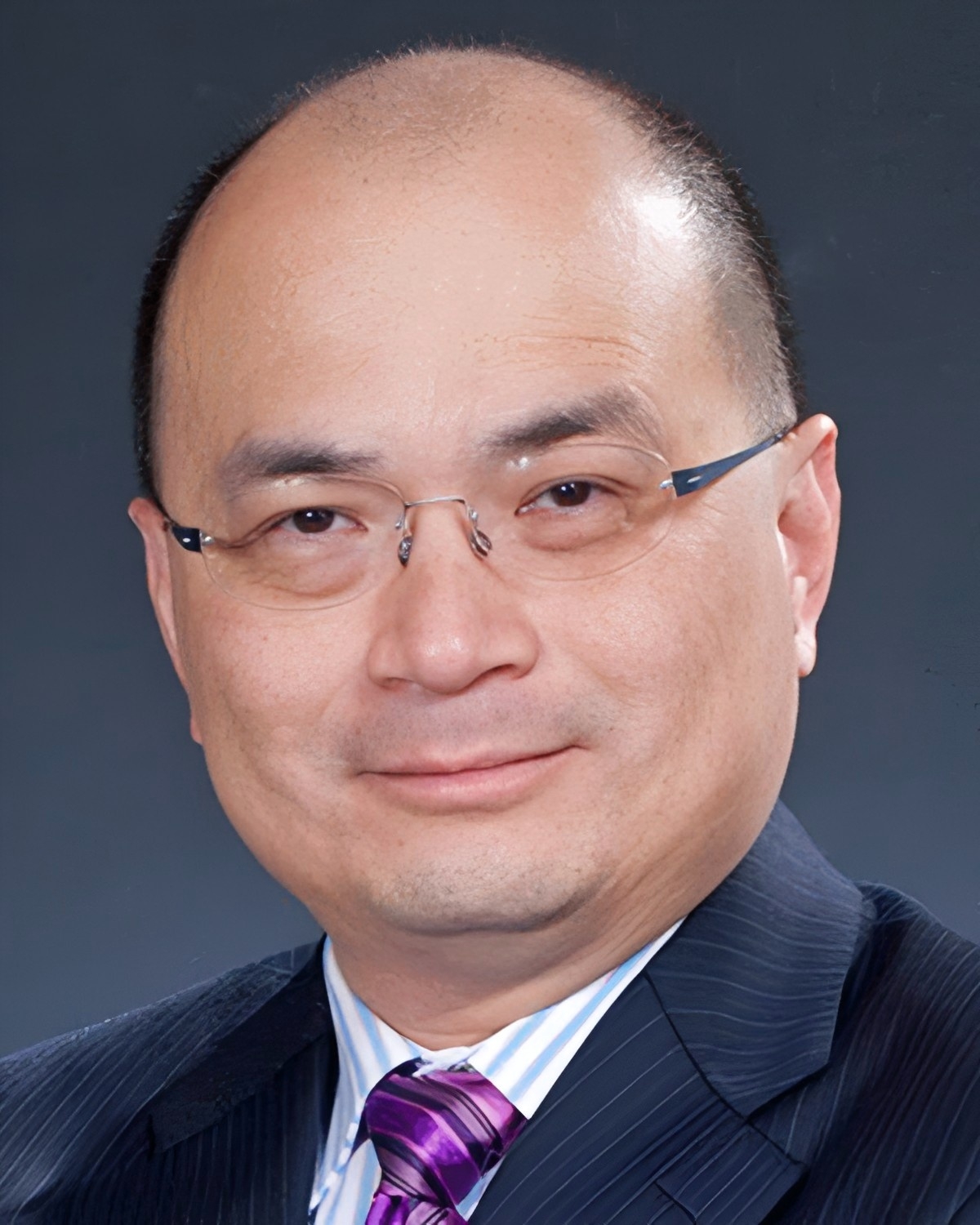}}]{Shangling Jui}
is the chief AI scientist for Huawei Kirin Chipset Solution. He is an expert in machine learning, deep learning, and artificial intelligence. Previously, he was the president of the SAP China Research Center and the SAP Korea Research Center. He was also the CTO of Pactera, leading innovation projects based on cloud and Big Data technologies. He received the Magnolia Award from the Municipal Government of Shanghai, in 2011.
\vspace{-30pt}
\end{IEEEbiography}


\begin{IEEEbiography}[{\includegraphics[width=1in,height=1.25in, clip,keepaspectratio]{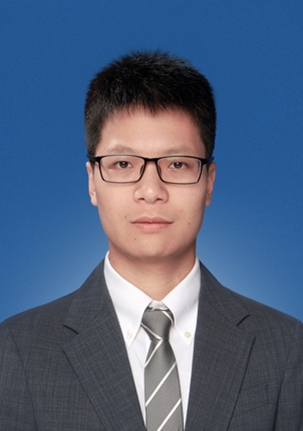}}]{Xiongkuo Min} received the B.E. degree from Wuhan University, Wuhan, China, in 2013, and the Ph.D. degree from Shanghai Jiao Tong University, Shanghai, China, in 2018, where he is currently a tenure-track Associate Professor with the Institute of Image Communication and Network Engineering. From Jan. 2016 to Jan. 2017, he was a visiting student at University of Waterloo. From Jun. 2018 to Sept. 2021, he was a Postdoc at Shanghai Jiao Tong University. From Jan. 2019 to Jan. 2021, he was a visiting Postdoc at The University of Texas at Austin and the University of Macau.  His research interests include image/video/audio quality assessment, quality of experience, visual attention modeling, extended reality, and multimodal signal processing. 
\vspace{-30pt}
\end{IEEEbiography}

\begin{IEEEbiography}[{\includegraphics[width=1in,height=1.25in,clip,keepaspectratio]{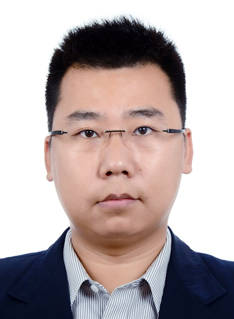}}]{Guangtao Zhai} received the B.E. and M.E. degrees from Shandong University, Shandong, China, in 2001 and 2004, respectively, and the Ph.D. degree from Shanghai Jiao Tong University, Shanghai, China, in 2009, where he is currently a Research Professor with the Institute of Image Communication and Information Processing.
From 2008 to 2009, he was a Visiting Student with the Department of Electrical and Computer Engineering, McMaster University, Hamilton, ON, Canada, where he was a Post-Doctoral Fellow from 2010 to 2012. From 2012 to 2013, he was a Humboldt Research Fellow with the Institute of Multimedia Communication and Signal Processing, Friedrich Alexander University of Erlangen-Nuremberg, Germany. He received the Award of National Excellent Ph.D. Thesis from the Ministry of Education of China in 2012. His research interests include multimedia signal processing and perceptual signal processing.
\vspace{-30pt}
\end{IEEEbiography}

\end{document}